\documentclass[conference]{IEEEtran}
\IEEEoverridecommandlockouts
\usepackage{cite}
\usepackage{amsmath,amssymb,amsfonts}
\usepackage{algorithmic}
\usepackage{graphicx}
\usepackage{textcomp}
\usepackage{xcolor}
\usepackage{graphicx}
\usepackage{subcaption}
\usepackage{amsmath}
\usepackage{amsthm}
\usepackage{booktabs}
\usepackage{algorithm}
\usepackage{algorithmic}
\usepackage{enumitem}
\usepackage{float}
\usepackage{multirow}
\usepackage{adjustbox}
\usepackage{caption} 
\usepackage{multirow}
\usepackage{scrextend}
\usepackage{multicol}
\usepackage{wrapfig}
\usepackage{breqn}
\usepackage{bm}
\usepackage{colortbl}
\usepackage{url}
\usepackage{caption}
\def\BibTeX{{\rm B\kern-.05em{\sc i\kern-.025em b}\kern-.08em
    T\kern-.1667em\lower.7ex\hbox{E}\kern-.125emX}}

\title{Continual Learning for Multivariate Time Series Tasks with Variable Input Dimensions\thanks{Accepted at ICDM'21 \copyright 2021 IEEE.}\\
}
\author{\IEEEauthorblockN{Vibhor Gupta, Jyoti Narwariya, Pankaj Malhotra, Lovekesh Vig, Gautam Shroff}
\IEEEauthorblockA{TCS Research,
New Delhi, India \\
\{g.vibhor, jyoti.narwariya, malhotra.pankaj, lovekesh.vig, gautam.shroff\}@tcs.com}
}
\begin{document}
\maketitle
\begin{abstract}
We consider a sequence of related multivariate time series learning tasks, such as predicting failures
for different instances of a machine from time series of multi-sensor data, or activity recognition tasks over different individuals from multiple wearable sensors. 
We focus on two under-explored practical challenges arising in such settings: (i) Each task may
have a different subset of sensors, i.e., providing different partial observations of the underlying `system'. 
This restriction can be due to different manufacturers in the former case, and people wearing more or less measurement devices in the latter (ii) We are not allowed to store or re-access data from a task once it has been observed at the task level. This may be due to privacy considerations in the case of people, or legal restrictions placed by machine owners. Nevertheless, we would like to (a) improve performance on subsequent tasks using experience from completed tasks as well as (b) continue to perform  better on past tasks, e.g., update the model and improve
predictions on even the first machine after learning from subsequently observed ones. 
We note that existing continual learning methods do not take into account variability in input dimensions arising due to different subsets of sensors being available across tasks, and struggle to adapt to such variable input dimensions (VID) tasks. 
In this work, we address this shortcoming of existing methods. To this end, we learn task-specific generative models and classifiers, and use these to augment data for target tasks. Since the input dimensions across tasks vary, we propose a novel conditioning module based on graph neural networks to aid a standard recurrent neural network. We evaluate the efficacy of the proposed approach on three publicly available datasets corresponding to two activity recognition tasks (classification) and one prognostics task (regression). We demonstrate that it is possible to significantly enhance the performance on future and previous tasks while learning continuously from VID tasks without storing data.
\end{abstract}

\begin{IEEEkeywords}
time series, continual learning, catastrophic forgetting, variable input dimensions, graph neural network.
\end{IEEEkeywords}

\section{Introduction}
With the increasing adoption of Internet of Things (IoT) \cite{da2014internet} technology, sensor data from various situations ranging from wearables to industrial machines is becoming increasingly available. Deep neural networks (DNNs) have been successfully applied on such data for forecasting, classification, anomaly detection, diagnostics, and prognostics \cite{p:lstm-ad,lai2018modeling,khan2018review}, \cite{wang2019deep},
with several applications in healthcare, utilities, etc.
Further, as shown in \cite{TIMENET, fawaz2018data}, a deep learning model trained on time-series across domains shows improved performance
on other domains. The potential for transfer across tasks is expected to be even more likely when tasks are similar, e.g., different instances of the same underlying dynamical system, e.g. humans in the case of activity recognition, or similar machines (engines/turbines, etc.) in the case of equipment health monitoring. 
In the extreme case, we might even like to consider data from all such instances as one dataset.

In the existing approaches, we face two important challenges in learning across a sequence of such multi-sensor tasks: (i) The number of sensors
used in each instance often differ, e.g., different people may be using different subsets of equipment (e.g., wrist-bands, phones, sensor-equipped shoes, etc.), and at different times. Similarly, machines installed in different factories, vehicles, or OEMs may be installed with a smaller or larger set of sensors often (e.g., temperature, vibration, pressure, etc.). However, most deep learning models typically assume fixed dimensional inputs and can handle variable temporal dimensions, i.e. variable length of the time series, but fail to adapt and generalize to variability in the number of available dimensions in multivariate
time series, where one or more dimensions are completely missing. Further, it may often be the case that some patterns (e.g., activities) cannot be detected using fewer sensors, and so tasks may also differ in the number of classes being encountered. 
(ii) While it is becoming increasingly common for artificial intelligence (AI) software providers to retain the right to continuously improve their machine learning models as new data is encountered, legal restrictions usually prohibit the storage and use of customer data beyond a certain limit, e.g. for privacy reasons
in the case of human activities, and even in the case of industrial data, the data used to train AI software legally belongs to the customer
and cannot be stored and used by the software vendor forever. Since tasks are encountered sequentially over time, a software vendor cannot use standard training techniques requiring multiple passes over all datasets to improve their models over time.

Motivated by the above, we consider a setting where multiple tasks with scarce and partially-observed sensor data, i.e., {\bf variable input dimensions} and {\bf classes} arrive sequentially. We would like to continuously improve a deep learning model as tasks are encountered, {\bf without being able to access data from previous tasks}, so as to {\it improve performance on subsequent tasks as well as on previous tasks}. (The latter requirement is actually important in practice as it is the rationale used by AI software vendors to claim the right to improve their models using customers' data, i.e., that this will enable future versions of the software to automatically perform better on their original task.) Note that this setting is related to the {\it online} continual learning problem \cite{kirkpatrick2017overcoming}, as also class-incremental learning \cite{kundu2020universal} where (in the case of classification) tasks may have varying subsets of classes. However, such approaches do not typically consider the variability in input dimensions, which is the main focus of this work.

Our key contributions can be summarized as follows:
\begin{itemize}[leftmargin=*]
	\item We motivate and formulate a novel and practically important problem of {\bf continual learning} from partially-observed multi-sensor (multivariate time-series) data, with a focus on the need to adapt to {\bf varying input dimensions (VID)} and classes across tasks with {\bf data-access prohibited across tasks}.
	\item We propose a novel modularized neural network architecture to handle variable input dimensions, with two main modules: 
	i. a {\bf core dynamics module} comprising an RNN that models the underlying dynamics of the system, 
	ii. a {\bf conditioning module using a graph neural network} (GNN \cite{battaglia2018relational}) that adjusts the activations of the core dynamics module for each time-series based on the combination of sensors available, effectively exhibiting different behavior depending on the available sensors.
	\item To leverage partially-observed sensor data, we build upon the idea of generative replay \cite{shin2017continual}, where task-specific generators generate variable dimensional time-series for subsequent processing by the task-solving core dynamics and conditioning modules. 
	\item We present an extensive empirical evaluation on diverse tasks from three real-world datasets to prove the advantage of our approach in contrast to vanilla architectures in terms of the ability to adapt to variable input dimensions and changing classes.	
\end{itemize}
\begin{figure*}[h]
	\centering
	\includegraphics[scale=0.17,trim={3cm 22.25cm 0cm 14.5cm}, clip]{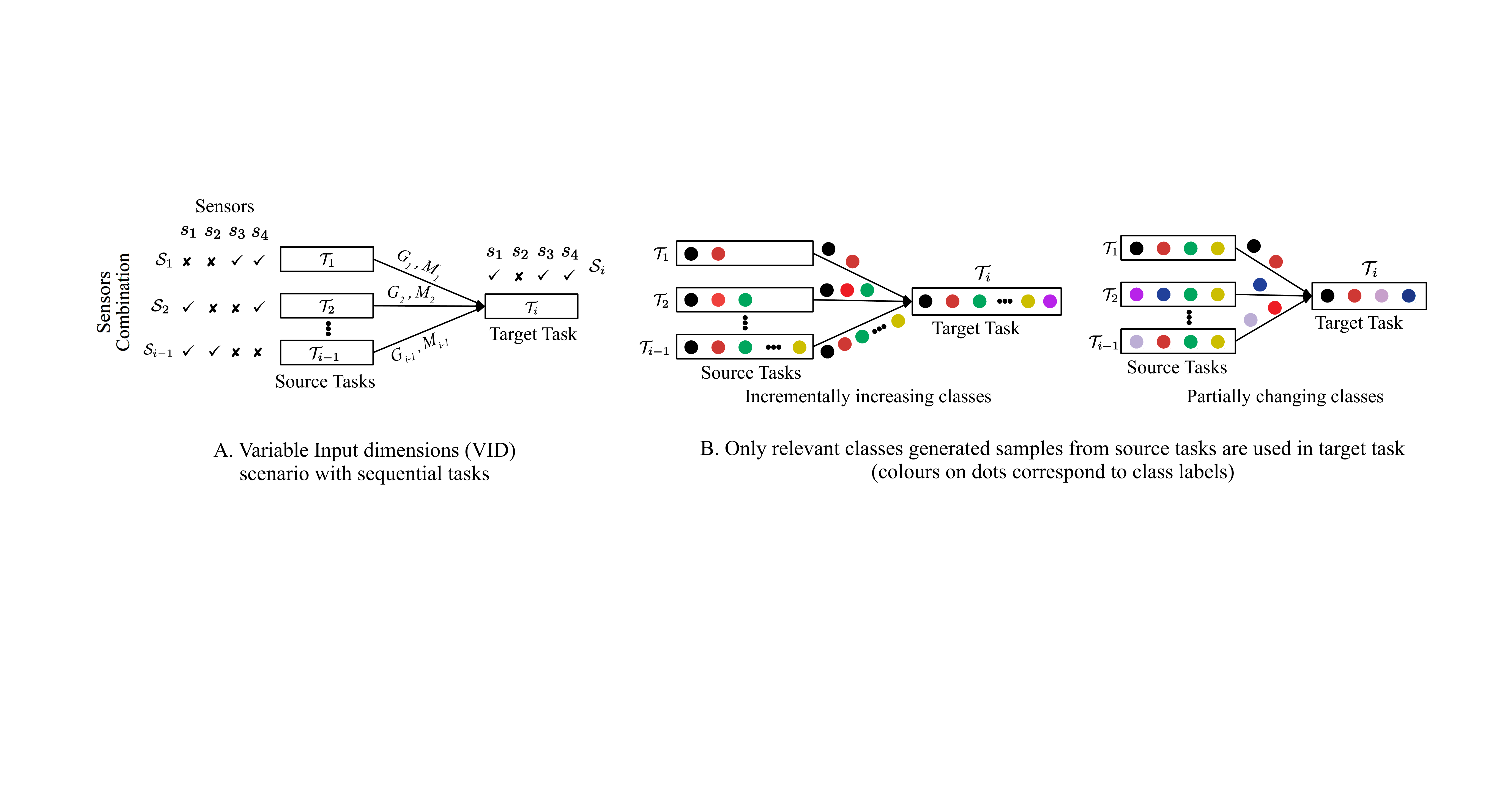}
	\caption{Scenarios considered (A and B). \label{fig:approach}}
\end{figure*}

\section{Problem Definition\label{sec:prob}}
Consider $m$ tasks $\mathcal{T} = (\mathcal{T}_{1}, \cdots, \mathcal{T}_{m})$
arriving sequentially. 
A task $\mathcal{T}_{i}$ ($i = 1, 2, \cdots, m)$, consists of $N$ instances of labeled multi-sensor data $\mathcal{D}_i = \{(\mathbf{x}_{i,r},y_{i,r})\}_{r=1}^N$, each time-series $\mathbf{x}_{i,r} \in \mathcal{X}_i$ with an associated target $y_{i,r} \in \mathcal{Y}_i$. The target space $\mathcal{Y}_i$ corresponds to class labels in case of classification tasks, or real-valued numbers (or vector) in case of regression tasks.
The goal in each task is to learn a \textit{solver model} ${M}_{i}$ with trainable parameters $\boldsymbol{ \theta}_i$ such that $M_i\!\!:\mathcal{X}_i \!\rightarrow \mathcal{Y}_i$, with estimated target $\hat{y}_{i, r} = {M}_{i}(\mathbf{x}_{i, r}; \boldsymbol{\theta}_{i})$.
Any task $\mathcal{T}_i$ is not allowed access to data from previous tasks $\mathcal{T}_j$ ($j={1,\ldots,i-1}$), but can leverage relevant generated samples obtained using generative models of these previous tasks.

We consider the following partially-observed multi-sensor tasks under the above-mentioned basic setup: 
i. tasks have a \textit{variable input space}, as illustrated in Fig. \ref{fig:approach}A, i.e. the input dimensions or available sensors change across tasks, i.e. for $r$-th time-series $\mathbf{x}_{i,r} = \{\mathbf{x}_{i,r}^t\}_{t=1}^{T_i}$ of length $T_i$ with $\mathbf{x}_{i,r}^t \in \mathbb{R}^{d_i}$, $ d_i \in \{1,2,\ldots,d\}$ is the number of available dimensions or sensors out of a maximum of $d$ possible dimensions. We refer to the set of available dimensions in $\mathcal{T}_i$ as $\mathcal{S}_i \subseteq \mathcal{S} = \{s_1,s_2,\ldots,s_d\}$, where $\mathcal{S}$ is the set of all the possible sensors.
ii. \textit{variable target space}: where classes of interest can vary across tasks, as illustrated in Fig. \ref{fig:approach}B; this is applicable for classification tasks only (but not for regression tasks).
The variable target space can correspond to either i. \textit{incrementally increasing classes}, where new classes are added for each new task while all old classes persist, or ii. a subset of old classes are present in the new task along with some new classes, i.e. the \textit{partially changing classes}.

\section{Related Work}
\subsection{Handling variable-dimensional time-series}
Several approaches in literature deal with variations in input data along temporal or spatial dimensions, i.e. variable length of time-series, or variable image sizes, e.g. via recurrent neural networks (RNNs) \cite{cho2014learning} for the time dimension, and variants of pooling operations for image applications \cite{he2015spatial}. 
Several approaches for handling missing values in multi-sensor data via neural networks have been proposed. 
For instance, \cite{che2018recurrent} study missing value problem in multi-sensor data by proposing a variant of the gated recurrent units (GRUs) \cite{cho2014learning} using knowledge of which dimensions of the input are missing and for how long. 
However, handling VID is a more difficult problem compared to the missing value problem in time-series.
Approaches dealing with missing values in time-series are not directly applicable in the VID setting where one or more dimensions of the time-series are completely missing (i.e. the missing \%age for the dimension/sensor is 100\%). Since these approaches would typically rely on one or more past values to extrapolate, and do not consider scenarios where a dimension is completely missing from first to last time step.

The generalization ability of GNNs across varying nodes has been studied in, e.g. GRU-CM \cite{gupta2020handling} and NerveNet \cite{wang2018nervenet} use the inherent ability of GNNs to generalize well on unseen combinations of test instances that are different from training instances. In GNMR \cite{narwariya2020graph}, GNNs leverage the knowledge of readily available graph structure to process multi-sensor time series data for the Remaining  Useful  Life (RUL) estimation in equipment health monitoring. However, these approaches neither take into account data-access restrictions across tasks nor variability in target spaces. 

Modular Universal Reparameterization (MUiR) \cite{meyerson2019modular} attempts to learn core neural modules which can be transferred across varying input and output dimensions. 
It relies on learning the core module by solving several sets of architecture-task problems with varying input and output dimensions.
However, MUiR does not study multi-sensor data setting, and relies on solving several tasks to learn the core neural module. Instead, our approach relies on a conditioning vector obtained via GNNs to allow adaptability to VID, and addresses additional constraints on simultaneous task-access as required by MUiR.
Recently, a neuro-evolutionary approach has been proposed in \cite{elsaid2020neuro} which studies the problem of structure-adaptive for time-series prediction. It relies on a series of mutation operations and crossover operations over the neural units. Instead of the computationally expensive neuro-evolutionary approaches, we take a different perspective on the problem of adapting to varying dimensionality where graph neural networks (GNNs) are used.

The recently proposed CondConv \cite{yang2019condconv} is similar in spirit to our work, i.e. it attempts to dynamically adapt the parameters of the neural network conditioned on the current input. While it focuses on adapting to each input with same dimensionality, our work focuses on adapting to inputs of varying dimensionality. 
Though significantly different in implementation and the end-objective, our approach also draws inspiration from such works, including \cite{rosenbaum2019routing,andreas2016neural}, where the parameters of the core neural network are dynamically adjusted as per the input.

\subsection{Continual Learning}
We review three recent approaches for continual learning. i. Experience replay: This approach maintains a storage buffer for rehearsing all previous tasks, e.g. \cite{rolnick2018experience,isele2018selective} and uses them along with new task data to avoid catastrophic forgetting. Our work is orthogonal to this approach, as real data-access prohibited across tasks. ii. Generative replay: Instead of storing real previous tasks data, a generative model can be used to mimic the real tasks distribution, e.g. \cite{shin2017continual,velik2014brain}. The generated samples of all previous tasks are used along with current tasks to adapt on old tasks. iii. Task-specific model: Instead of storing or replaying previous tasks data, different model components such as expanding model for each new task \cite{yoon2017lifelong} or using supermask to select task-specific subnetworks, while keeping the weights of model \cite{wortsman2020supermasks}  fixed to avoids catastrophic forgetting. 
However, these approaches focus on variable classes across tasks, and do not usually consider the VID setting. In fact, as we show empirically in Section \ref{sec:exp}, these approaches struggle in the VID setting.

\section{Approach\label{sec:approach}}
\begin{figure}[h!]
	\centering
	\includegraphics[scale=0.2,trim={35cm 25.25cm 31cm 12.0cm}, clip]{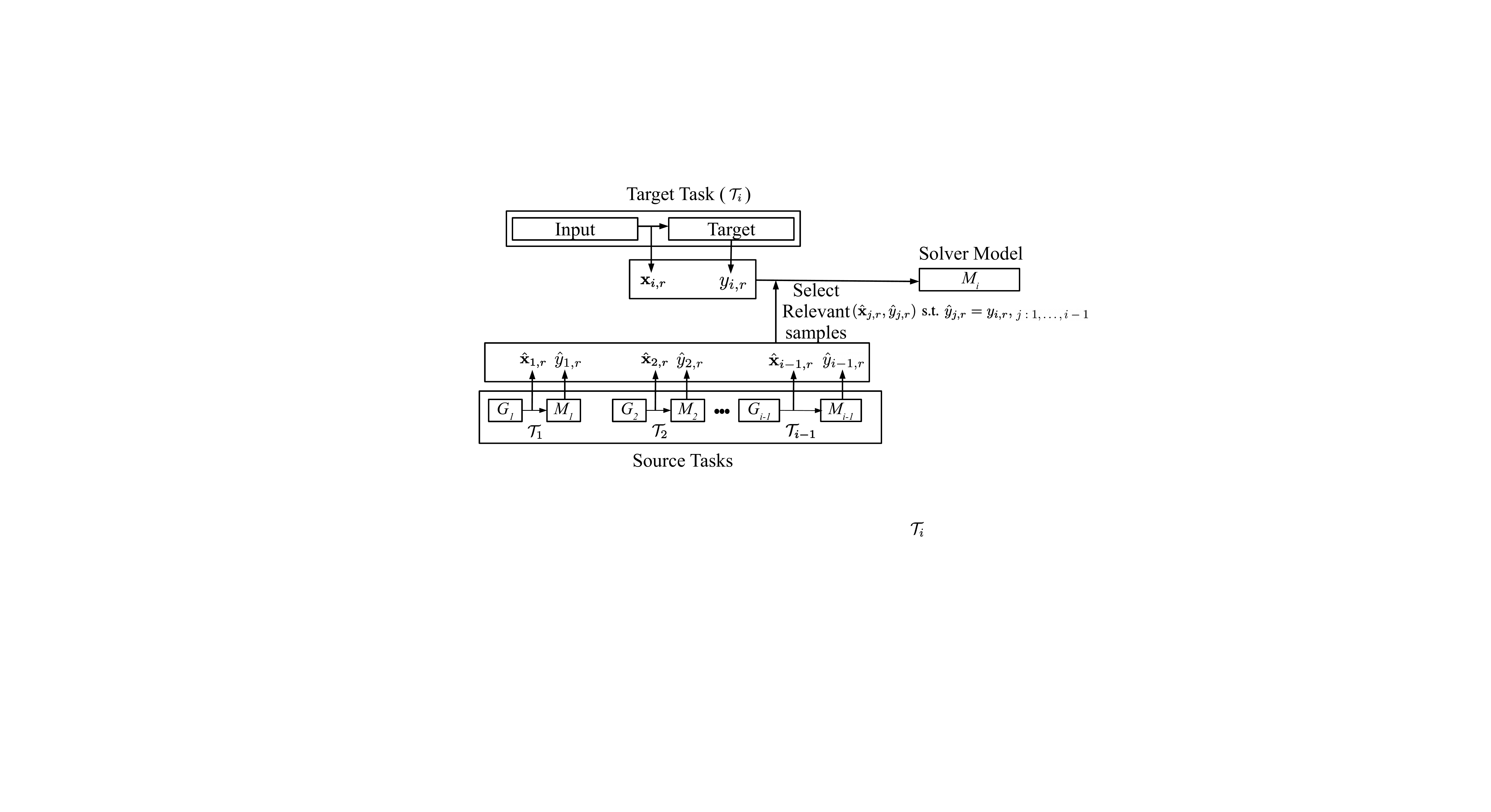}
	\caption{Using relevant generator and solver for augmenting training data for task ($\mathcal{T}_i$). \label{fig:relevant_samples}}
\end{figure}

Our approach consists of learning a series of generator and solver models, one for each task, as shown in Fig. \ref{fig:approach}A and \ref{fig:relevant_samples}. 
For any task $\mathcal{T}_i$, a solver model ${M}_i$ is trained in a supervised manner using labeled data, a generator model ${G}_i$ is trained in an unsupervised manner as a generative model for the data $\mathcal{D}_i$ from $\mathcal{T}_i$.
We denote the original and generated data for $\mathcal{T}_i$ as $\mathcal{D}_i$ and $\hat{\mathcal{D}}_i$, respectively.
While ${G}_i$ generates unlabeled time-series data, the corresponding labels are estimated using the learned ${M}_i$. 
This labeled generated data $\hat{\mathcal{D}}_i$ serves as additional training data for subsequent data-scarce tasks $\mathcal{T}_j$ ($j=i+1,\ldots,m$) instead of $\mathcal{D}_i$.
\begin{figure*}[h!]
	\centering
	\includegraphics[scale=0.215,trim={22cm 18.25cm 25cm 8cm}, clip]{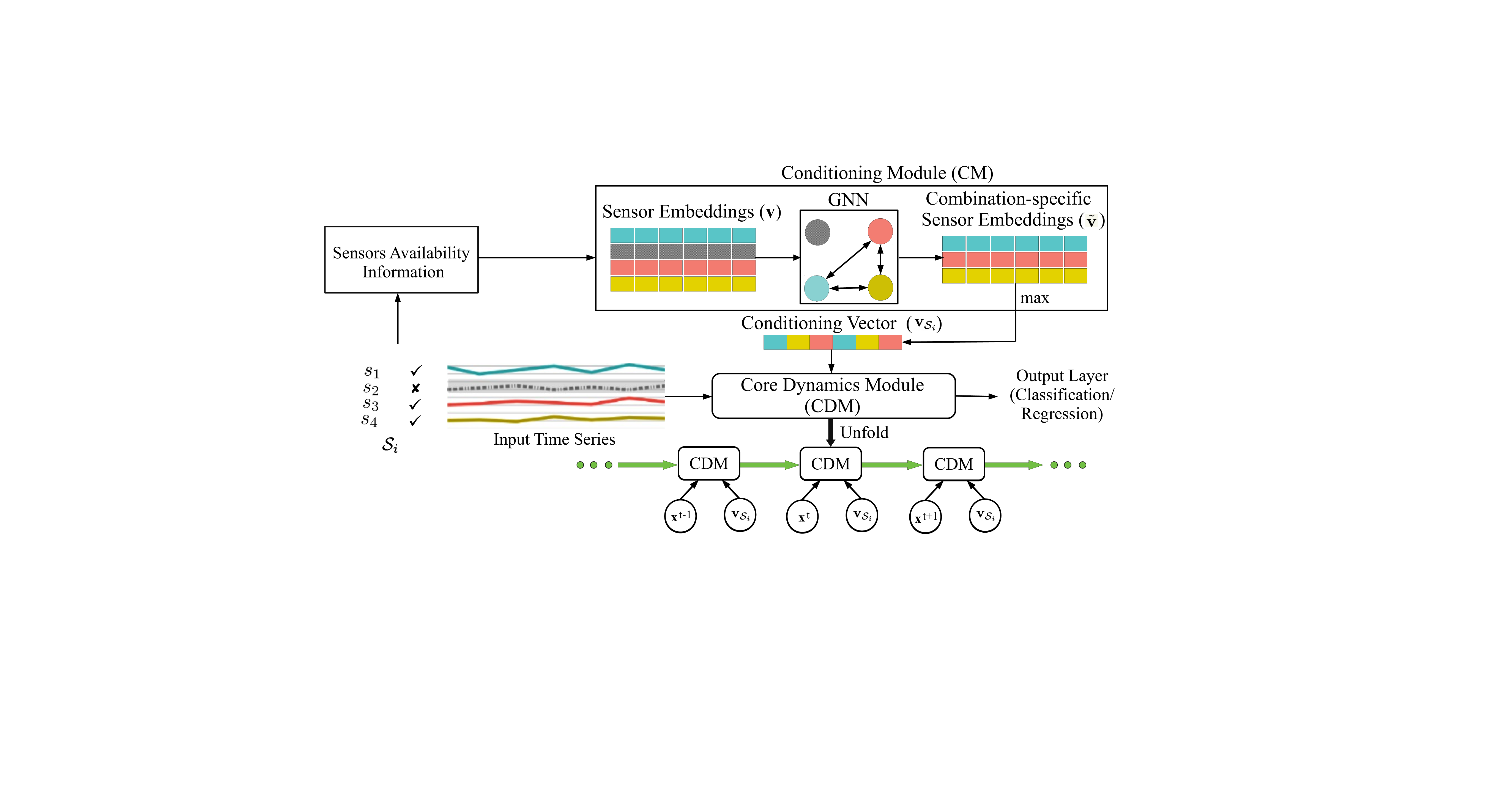}
	\caption{Flow-diagram of the Solver-CM model using GNN-based Conditioning Module. The available sensors in a time series and corresponding active embeddings and nodes are shown in color while the missing sensor/dimension is shown in grey color (dashed). The active nodes corresponding to the available sensors exchange messages with each other via a graph neural network, and generate a conditioning vector that is used by the core dynamics module (a recurrent neural network) processing the time series to estimate the target. \label{fig:solver_CM}}
\end{figure*}
\subsection{Generator Models}
Our approach using generated data for subsequent tasks is similar in spirit to existing work in generative replay (GR) for continual learning, e.g. \cite{shin2017continual,achille2018life}. 
However, instead of maintaining a common generator (CG) across tasks, we maintain a separate independent generator (IG) ${G}_i$ for each task. We refer prior work such as  \cite{shin2017continual} relying on CG as \textbf{GR-CG}, and our proposed approach relying on IG as \textbf{GR-IG}.

This design choice of having an GR-IG per task is crucial to handle VID across tasks: when dimensions across tasks vary, a GR-CG with fixed input dimensions $d$ cannot easily adapt to the missing $d-d_i$ dimensions in $\mathcal{T}_i$ as there is no training signal for those dimensions. Even if the GR-CG can model the $\mathcal{S}_i$ sensors of the current task $\mathcal{T}_i$ well, but it fails to adapt to a new combination of sensors in $\mathcal{S}_{i+1}$ from the task $\mathcal{T}_{i+1}$.
In contrast, while training the GR-IG ${G}_i$, the input dimensions correspond only to the original relevant sensors $\mathcal{S}_i$ for that task, and the burden of handling VID is passed on to the solver model ${M}_i$ whose novel modular structure relying on GNNs is specifically designed to be able to handle VID data. 
We analyze and validate the advantage of GR-IG over GR-CG empirically in Section \ref{sec:exp}.

In this work, we use variational autoencoders (VAEs) \cite{kingma2013auto,achille2018life} as generative models for the data distribution $p(\mathcal{D})$. We use VAEs for ease of training. However, the proposed approach is agnostic to the choice of generator, and other generative models such as generative adversarial networks (GANs, as used in \cite{shin2017continual}), are equally applicable. The encoder and decoder of VAEs are RNNs with GRUs that allow for capturing the temporal dependencies in multi-sensor data. The loss function is the same as in standard VAEs (refer \cite{kingma2013auto} and Appendix for more details).

\subsection{Solver Model}
The training data $\tilde{\mathcal{D}}_i$ for the solver model ${M}_{i}$ consists of original data $\mathcal{D}_i$ from $\mathcal{T}_i$ as well as relevant\footnote{We also consider the scenario of using all classes instead of relevant classes samples from the previous generated tasks data, but found using the relevant classes samples performs better, especially in partially changing classes scenario.} generated data $\bigcup_{j=1}^{i-1}\hat{\mathcal{D}_{j}}$ from the previous tasks $\mathcal{T}_1,\ldots,\mathcal{T}_{i-1}$. This inevitably results in VID data if different tasks have different dimensions. 
However, the training data across tasks can have VID as well as variable target space. VID is handled via solver model, whereas variable target space is handled by using only those generated samples from the previous tasks $\mathcal{T}_1,\ldots,\mathcal{T}_{i-1}$ for which the estimated class labels correspond to classes in $\mathcal{T}_i$, as illustrated in Fig. \ref{fig:approach}B. 
Any point $\mathbf{x}^t$ at a timestep $t$ in time-series $\mathbf{x} \in \tilde{\mathcal{D}}_i = \bigcup_{j=1}^{i-1}\hat{\mathcal{D}_{j}} \cup \mathcal{D}_i$, is mapped to a vector in $\mathbb{R}^d$ with a fixed consistent indexing on the sensors in $\mathcal{S}$, irrespective of the number of available sensors in $\mathbf{x}$. The unavailable sensors are mean-imputed, i.e. the mean value of the sensor across other instances where it is available is used.

The solver model consists of the two modules, as illustrated in Fig. \ref{fig:solver_CM}:
	i. \textit{core dynamics module} (CDM), which is a standard GRU network that learns the dynamics of the system and ingests a fixed-dimensional time-series where the missing dimensions are imputed with a constant (mean) value, 
	ii. \textit{conditioning module} (CM), which generates a ``conditioning vector'' as a function of the set of available sensors irrespective of the readings those sensors take for a particular time-series instance as shown in Appendix Fig. \ref{fig:Conditioning module}. 
	This conditioning vector is passed as an additional input to the CDM allowing it to adjust its internal computations and activations according to the combination of available sensors.
	The conditioning vector is in-turn obtained from ``sensor embedding vectors'' via a GNN.
	Note that the CDM and the CM along with the sensor embeddings are trained in an end-to-end fashion via stochastic gradient descent (SGD). We next provide architecture details of the two modules.

\subsubsection{Conditioning Module (CM)}
The CM based on GNNs is one of the crucial modules in the overall architecture as it allows to handle missing dimensions (variables) in multivariate time series. For this, we consider a fully-connected bi-directional graph where each node corresponds to a sensor, such that each active node (corresponding to available sensors in a time series) in the graph is connected to every other active node.  
Any node in the graph is updated by using aggregated messages from its neighboring nodes. The conditioning vector is obtained as a summary of the messages exchanged between the active nodes. This conditioning vector guides the core dynamics module (CDM) about  the current ``context'', i.e. available sensors, so that the computations of the CDM can be adjusted according to the available sensors. 

More specifically, corresponding to the set of sensors $\mathcal{S}$, consider a graph $G(\mathcal{V},\mathcal{E})$ with nodes or vertices $\mathcal{V}$ and edges $\mathcal{E}$, with one node $v_s \in \mathcal{V}$ for every $s \in \mathcal{S}$ such that $|\mathcal{V}|=|\mathcal{S}|$.
The set of neighboring nodes for any node $v_s$ is denoted by $\mathcal{N}_{G}(v_s)$.
Each sensor $s \in \mathcal{S}$ is associated with a learnable embedding vector $\mathbf{v}_s \in \mathbb{R}^{d'}$, where $d'$ is a hyperparameter. 
While training a model for $\mathcal{T}_i$, the possible dimensions or sensor combinations can be any of $\mathcal{S}_1, \mathcal{S}_2,\ldots,\mathcal{S}_i$, corresponding to the sensor combinations present in $\hat{\mathcal{D}}_1, \hat{\mathcal{D}}_2, \ldots,\hat{\mathcal{D}}_{i-1}$ or $\mathcal{D}_i$. For a given set or combination of sensors $\mathcal{S}_j$ ($j=1,\ldots, i$), only the corresponding nodes $\mathcal{V}_j \subseteq \mathcal{V}$ are considered to be \textit{active}, and contribute to obtaining the combination-specific conditioning vector $\mathbf{v}_{\mathcal{S}_j}$. 
For the active nodes, the graph is assumed to be \textit{fully-connected} such that each active node in the graph is connected to every other active node. 
Any edge is active only if both the participating nodes are active.

The GNN corresponding to this graph consists of a node-specific feed-forward network $f_n$ and an edge-specific feed-forward network $f_e$; $f_n$ and $f_e$ are shared across the nodes and edges in the graph, respectively.
For any active node $v_a \in \mathcal{V}_i$, the node vector $\mathbf{v}_a$ is updated using the GNN as follows: 
\begin{align}
	\mathbf{v}_{al} &= f_{e}([\mathbf{v}_{a},\mathbf{v}_{l}]; \boldsymbol{\theta}_{i}^{e}) \label{eq1}, \quad \forall v_l \in \mathcal{N}_{G}(v_a),\\
	\tilde{\mathbf{v}}_{a} &= f_{n}([\mathbf{v}_{a},\sum_{\forall l} \mathbf{v}_{al}]; \boldsymbol{\theta}_{i}^{n}) \label{eq2},
\end{align}
where $f_{e}$ and $f_{n}$ both consist of feedforward layers with Leaky ReLUs \cite{maas2013rectifier} and dropout \cite{srivastava2014dropout}, and with learnable parameters $ \boldsymbol{\theta}_{i}^{e}$ and $ \boldsymbol{\theta}_{i}^{n}$, respectively. 
While $f_{e}$ computes the message from node $v_l$ to $v_a$, $f_{n}$ updates the node vector $\mathbf{v}_a$ to $\tilde{\mathbf{v}}_a$ using messages from its neighboring nodes (Eqn. \ref{eq2}). 
Finally, the conditioning vector $\mathbf{v}_{\mathcal{S}_j}$ specific to the set of sensors $\mathcal{S}_j$ is obtained from the updated node vectors as
\begin{equation}\label{eq3}
	\mathbf{v}_{\mathcal{S}_j} = \mathtt{max}(\{\tilde{\mathbf{v}}_a\}_{v_a \in \mathcal{V}_j}), 
\end{equation}
where $\mathtt{max}$ returns the dimension-wise maximum value\footnote{We also tried averaging instead of max operation to summarize the vectors of available sensors, but found max to work better.} across the updated node vectors.
It is noteworthy that the summation over the messages across nodes in Eqn. \ref{eq2} and the $\mathtt{max}$ operation in Eqn. \ref{eq3} essentially provide the desired ability to process VID. 

\subsubsection{Core Dynamics Module (CDM)}
As mentioned earlier, any time-series $\mathbf{x}_{i, r} \in \tilde{\mathcal{D}}_i$ is first converted to the $d$-dimensional time-series $\tilde{\mathbf{x}}_{i, r}$ with mean-imputation for the unavailable sensors.
This time-series along with its conditioning vector, say $\mathbf{v}_{\mathcal{S}_j}$ (assuming this time-series comes from generated data $\hat{\mathcal{D}}_j$) for the task $\mathcal{T}_{i}$ are processed by the CDM as follows:
\begin{align}
	\mathbf{z}^t_{i, r} &= GRU([\tilde{\mathbf{x}}_{i, r}^t, \mathbf{v}_{\mathcal{S}_j}],\mathbf{z}^{t-1}_{i, r}; \boldsymbol{\theta}_{i}^{GRU}), \quad t:1,\ldots,T_i \label{eq6}\\
	\hat{y}_{i, r} &=f_{o}(\mathbf{z}^{T_i}_{i, r};\boldsymbol{\theta}_{i}^{o})\label{eq7},
\end{align}
where $GRU$ is a multi-layered RNN having $\boldsymbol{\theta}_{i}^{GRU}$ learnable parameters that gives feature vector $\mathbf{z}^{T_i}$ at the last timestep $T_i$, and $\mathbf{v}_{\mathcal{S}_j}$ is concatenated with $d$-dimensional sensor values at each time-step. 
At last, the estimate $\hat{y}_{i, r}$ for $y_{i, r}$ is obtained via $f_{o}$ consisting of ReLU layer(s) followed by softmax or sigmoid layer corresponding to classification or regression, respectively, with $\hat{y}_{i, r} \in [0,1]^K$ for classification, and $\hat{y}_{i, r} \in \mathbb{R}$ for regression. For the  RUL estimation regression task, we use min-max normalized targets such that they lie in $[0,1]$.
All the steps described above from the Eqn. \ref{eq1} to Eqn. \ref{eq7} can be consolidated as follows:
 $\hat{y}_{i,r} = {M}_{i}(\mathbf{x}_{i,r}; \boldsymbol{\theta}_{i})$,
where $\boldsymbol{\theta}_{i}$ includes all the parameters $(\boldsymbol{\theta}_{i}^{e}, \boldsymbol{\theta}_{i}^{n}, \boldsymbol{\theta}_{i}^{GRU}$ and $ \boldsymbol{ \theta}_{i}^{o})$ of $M_{i}$.

\section{Training Objective}
Denoting the training objectives for classification (standard cross-entropy loss) or regression (squared-error loss) as $\mathcal{L}(y_{i, r}, \hat{y}_{i, r}; \boldsymbol{\theta}_{i})$, we have:
\begin{dmath}
	\mathcal{L}_i= \frac{q}{N}\sum_{r = 1}^{N}\mathcal{L}\big(y_{i, r}, \hat{y}_{i, r}; \boldsymbol{\theta}_{i}\big) + \Big(\frac{1 - q}{(i-1)N}\Big)\sum_{j = 1}^{i-1}\sum_{r=1}^{N}\mathcal{L}\Big(y_{j,r}, {M}_{i}(\mathbf{\hat{x}}_{j,r}; \boldsymbol{\theta}_i); \boldsymbol{\theta}_{i}\Big), 
\end{dmath}
where, $0 < q \leq 1$ is the importance given to data from $\mathcal{T}_i$ and $N$ is the number of instances. Higher values of $q$ imply more weightage to samples from $D_i$. In this work, we use equal weightage to all samples, i.e. $q=0.5$. Note that $y_{j,r}$ is the label corresponding to class with maximum probability as given by the solver model ${M}_{j}$ for the generated sample $\mathbf{\hat{x}}_{j,r}$.

\section{Evaluation\label{sec:exp}}

\subsection{Experimental Setup}
We consider five tasks arriving sequentially with each sequential task containing a small amount of labeled training data (only 5$\%$ of training instances in case of DSADS \cite{altun2010human} and HAR \cite{anguita2012human}, and 10 engines data in Turbofan \cite{saxena2008damage} - FD001 ). The training data for each task undergoes an 80-20 split into a train and validation set, where the validation set is used for early stopping.
Apart from evaluating for fixed input dimensions (FID) and fixed classes across tasks, we consider: i. variable input dimensions (VID) across the tasks for all three datasets, and ii. variable target space across the tasks for the classification datasets.
We randomly remove 40\% of sensors in each task for VID scenario. 
For \textit{incrementally increasing classes} scenario, one class is added for each new task while all the old classes persist in DSADS and HAR. For the \textit{partially changing classes} scenario, two and one old class(es) are replaced with two and one new class(es) for each new task in case of DSADS and HAR, respectively.

For DSADS and HAR datasets, we use sensor-wise z-normalization for each input dimension/sensor, whereas sensor-wise min-max normalization is used for the Turbofan dataset. We used classification error rates and root mean squared error (RMSE) as the performance metrics for the classification and regression, respectively. For each scenario, we consider five different sets\footnote{In variable target space, each set has different combination of classes.} of five sequentially arriving tasks. In Table \ref{tab:results}, we report the average value across five sets.  Refer Appendix for further details on datasets and hyperparameters.

\subsection{Approaches Considered}
For comparison, we consider the following approaches:
\begin{table*}[h]
	\caption{Performance comparison of various approaches on sequential tasks. *Note: UB with standard solver for VID scenario does not serve as an upper bound despite access to all past data due to difficulty in handling VID. Bold implies approach is statistically significance against others with $p < 0.05$. \textit{Our proposed approach GR-IG (ours) has significant gain compared to other approaches on most of the tasks, in both the scenarios FID and VID.}}
	\centering
	\scalebox{0.8}{
		\begin{tabular}{|c|c|c|cccccc||cccccc|ccccc|}
			\hline
			&&\multicolumn{7}{c||}{\textbf{Fixed Input Dimensions (FID) }}& \multicolumn{11}{c|}{\textbf{Variable Input Dimensions (VID)}} \\
			\cline{3-20}
			\multirow{4}{*}{\begin{tabular}[c]{@{}c@{}}\textbf{Dataset}\\\\\\\end{tabular}}&
			\multirow{4}{*}{\begin{tabular}[c]{@{}c@{}}\textbf{Scenario}\\\\\\\end{tabular}}&
			\multirow{4}{*}{\begin{tabular}[c]{@{}c@{}}\textbf{Task}\\\textbf{ID}\\\\\end{tabular}} & \multicolumn{6}{c||}{\textbf{Standard Solver}} & \multicolumn{6}{c|}{\textbf{Standard Solver}} & \multicolumn{5}{c|}{\textbf{Solver-CM }} \\
			\cline{4-20}
			&&&\multicolumn{1}{l|}{\multirow{2}{*}{\begin{tabular}[c|]{@{}c@{}}\textbf{TS}\\\end{tabular}}}&\multicolumn{1}{l|}{\multirow{2}{*}{\begin{tabular}[c|]{@{}c@{}}\textbf{FT}\\\end{tabular}}}&\multirow{2}{*}{\begin{tabular}[c|]{@{}c@{}}\textbf{SMT}\\\end{tabular}}&\multicolumn{2}{|c|}{\textbf{GR}} &\multirow{2}{*}{\begin{tabular}[c]{@{}c@{}}\cellcolor[gray]{0.7}\textbf{UB}\\\end{tabular}}&\multicolumn{1}{l|}{\multirow{2}{*}{\begin{tabular}[c|]{@{}c@{}}\textbf{TS}\\\end{tabular}}}&\multicolumn{1}{l|}{\multirow{2}{*}{\begin{tabular}[c|]{@{}c@{}}\textbf{FT}\\\end{tabular}}}&\multirow{2}{*}{\begin{tabular}[c|]{@{}c@{}}\textbf{SMT}\\\end{tabular}}&\multicolumn{2}{|c|}{\textbf{GR}} &\multirow{2}{*}{\begin{tabular}[c]{@{}c@{}}\cellcolor[gray]{0.7}\textbf{UB$^*$}\\\end{tabular}}&\multicolumn{1}{l|}{\multirow{2}{*}{\begin{tabular}[c|]{@{}c@{}}\textbf{FT}\\\end{tabular}}}&\multirow{2}{*}{\begin{tabular}[c|]{@{}c@{}}\textbf{SMT}\\\end{tabular}}&\multicolumn{2}{|c|}{\textbf{GR}} &\multirow{2}{*}{\begin{tabular}[c]{@{}c@{}}\cellcolor[gray]{0.7}\textbf{UB}\\\end{tabular}}\\
			\cline{7-8} \cline{13-14} \cline{18-19}
			&&&\multicolumn{1}{l|}{}     &\multicolumn{1}{l|}{}   &\   \textbf{\cite{wortsman2020supermasks}}& \multicolumn{1}{|l|}{   \textbf{\begin{tabular}[c]{@{}c@{}}\textbf{CG}\\\cite{shin2017continual}\end{tabular}}} & \multicolumn{1}{l|}{   \textbf{\begin{tabular}[c]{@{}c@{}}\textbf{IG}\\\textbf{(ours)}\end{tabular}}} &   &\multicolumn{1}{l|}{}     &\multicolumn{1}{l|}{}   &\   \textbf{\cite{wortsman2020supermasks}}& \multicolumn{1}{|l|}{   \textbf{\begin{tabular}[c]{@{}c@{}}\textbf{CG}\\\cite{shin2017continual}\end{tabular}}} & \multicolumn{1}{l|}{   \textbf{\begin{tabular}[c]{@{}c@{}}\textbf{IG}\\\textbf{(ours)}\end{tabular}}} & & \multicolumn{1}{l|}{}   &\   \textbf{\cite{wortsman2020supermasks}}& \multicolumn{1}{|l|}{   \textbf{\begin{tabular}[c]{@{}c@{}}\textbf{CG}\\\cite{shin2017continual}\end{tabular}}} & \multicolumn{1}{l|}{   \textbf{\begin{tabular}[c]{@{}c@{}}\textbf{IG}\\\textbf{(ours)}\end{tabular}}} & \\
			\hline
			
			\parbox[t]{2mm}{\multirow{15}{*}{\rotatebox[origin=c]{90}{\textbf{DSADS}}}} & 
			
			\multirow{5}{*}{\begin{tabular}[c]{@{}c@{}}Fixed\\ Classes\\\end{tabular}}

			&1 & 16.3 & 16.3 & 16.2 & 16.3 & 16.3 & \textit{16.3} & 27.8 & 27.8 & 29.2 & 27.8 & 27.8 & \textit{27.8} & 27.8 & 29.2 & 27.8 & 27.8 & \textit{27.8} \\
&&2&11.9 & 12.0 & 11.9 & 10.9 & \textbf{10.5} & \textit{9.0} & 29.4 & 30.4 & 27.3 & 29.9 & 28.6 & \textit{25.6} & 26.6 & 25.7 & 25.2 & \textbf{20.7} & \textit{13.9} \\
&&3&17.9 & 14.7 & 12.7 & 12.4 & \textbf{9.7} & \textit{6.5} & 29.6 & 26.2 & 28.3 & 36.5 & 25.1 & \textit{22.7} & 23.4 & 26.8 & 36.1 & \textbf{18.7} & \textit{11.3} \\
&&4&18.9 & 13.2 & 13.6 & 10.2 & \textbf{8.0} & \textit{4.5} & 26.1 & 26.9 & 25.1 & 58.3 & 23.6 & \textit{20.9} & 23.3 & 24.1 & 52.2 & \textbf{16.2} & \textit{7.7} \\
&&5&15.7 & 12.3 & 12.1 & 8.7 & \textbf{7.4} & \textit{4.0} & 26.1 & 24.1 & 25.3 & 54.7 & 22.4 & \textit{19.1} & 22.3 & 22.6 & 48.6 & \textbf{13.9} & \textit{7.0} \\
			\cline{2-20}
			
			&\multirow{5}{*}{\begin{tabular}[c]{@{}c@{}}Incrementally\\ Increasing\\ Classes\\\end{tabular}}
			
			&1 & 13.4 & 13.4 & 13.8 & 13.4 & 13.4 & \textit{13.4} & 19.1 & 19.1 & 17.9 & 19.1 & 19.1 & \textit{19.1} & 19.1 & 17.9 & 19.1 & 19.1 & \textit{19.1} \\
&&2&15.6 & 14.9 & 13.8 & 13.2 & \textbf{11.1} & \textit{9.7} & 21.4 & 26.9 & 22.9 & 24.4 & 22.8 & \textit{18.8} & 25.6 & 20.6 & 21.2 & \textbf{15.3} & \textit{12.7} \\
&&3&12.8 & 11.9 & 10.9 & 11.3 & \textbf{9.2} & \textit{6.9} & 15.7 & 34 & 19.5 & 39.4 & 26.3 & \textit{25.0} & 24.3 & 18.6 & 33.5 & \textbf{13.5} & \textit{10.8} \\
&&4&11.7 & 8.4 & 9.4 & 7.5 & \textbf{6.0} & \textit{4.3} & 21.2 & 35.7 & 26.1 & 45.3 & 30.7 & \textit{23.7} & 28.9 & 23.6 & 38.3 & \textbf{13.9} & \textit{9.5} \\
&&5&9.2 & 8.3 & 8.4 & \textbf{6.3} & \textbf{6.1} & \textit{3.8} & 13.4 & 39.6 & 24.8 & 47.6 & 36.3 & \textit{27.9} & 27.8 & 12.3 & 43.1 & \textbf{11.8} & \textit{6.7}\\
			\cline{2-20}
			
			&\multirow{5}{*}{\begin{tabular}[c]{@{}c@{}}Partially\\ Changing\\ Classes\\ \end{tabular}}
			&1 & 13.4 & 13.4 & 13.8 & 13.4 & 13.4 & \textit{13.4} & 19.1 & 19.1 & 17.9 & 19.1 & 19.1 & \textit{19.1} & 19.1 & 17.9 & 19.1 & 19.1 & \textit{19.1} \\
&&2&12.3 & 11.7 & 11.2 & 11.3 & \textbf{9.5} & \textit{8.0} & 16.5 & 29.9 & 20.5 & 26.7 & 23.1 & \textit{17.7} & 26.1 & 19.7 & 27.9 & \textbf{14.9} & \textit{13.1} \\
&&3&13.5 & 12.4 & 15.0 & 12.4 & \textbf{9.8} & \textit{7.4} & 18.5 & 25.4 & 19.4 & 32.4 & 25.4 & \textit{18.8} & 22.3 & 17.1 & 29.4 & \textbf{14.8} & \textit{13.4} \\
&&4&14.3 & 13.4 & 15.1 & 12.4 & \textbf{10.1} & \textit{7.2} & 19.3 & 23.7 & 23.3 & 41.2 & 22.0 & \textit{19.5} & 20.9 & 20.4 & 36.7 & \textbf{17.7} & \textit{13.9} \\
&&5&15.4 & 13.3 & 12.7 & 10.6 & \textbf{8.8} & \textit{6.5} & 18.6 & 23.4 & 19.7 & 30.5 & 21.7 & \textit{18.6} & 20.6 & 19.0 & 32.1 & \textbf{15.2} & \textit{11.6} \\
			\hline \hline
			
			\parbox[t]{2mm}{\multirow{15}{*}{\rotatebox[origin=c]{90}{\textbf{HAR}}}} &
			\multirow{5}{*}{\begin{tabular}[c]{@{}c@{}}Fixed \\ Classes\\\end{tabular}}

			&1 & 15.7 & 15.7 & 16.1 & 15.7 & 15.7 & \textit{15.7} & 30.2 & 30.2 & 30.9 & 30.2 & 30.2 & \textit{30.2} & 30.2 & 30.9 & 30.2 & 30.2 & \textit{30.2} \\
&&2&17.1 & 16.3 & 16.2 & 15.8 & \textbf{15.4} & \textit{14.4} & 41.5 & 46.9 & 41.2 & 46.9 & 40.1 & \textit{37.4} & 44.9 & 38.8 & 37.3 & \textbf{30.8} & \textit{27.7} \\
&&3&18.2 & 16.1 & 15.8 & 15.1 & \textbf{14.4} & \textit{12.8} & 34.4 & 42.7 & 33.2 & 52.1 & 34.0 & \textit{32.4} & 40.6 & 32.1 & 40.9 & \textbf{30.2} & \textit{26.2} \\
&&4&17.4 & 15.5 & 15.2 & \textbf{13.4} & 13.8 & \textit{11.9} & 34.6 & 45.7 & 32.2 & 72.2 & 31.4 & \textit{28.4} & 43.0 & \textbf{28.4} & 68.2 & \textbf{28.2} & \textit{23.6} \\
&&5&17.6 & 15.7 & 15.2 & 13.5 & \textbf{13.1} & \textit{11.4} & 35.0 & 31.3 & 32.1 & 72.9 & 29.0 & \textit{26.2} & 30.4 & 32.5 & 71.5 & \textbf{23.7} & \textit{20.1} \\
			\cline{2-20}
			&\multirow{5}{*}{\begin{tabular}[c]{@{}c@{}}Incrementally\\ Increasing\\ Classes\\\end{tabular}}
			&1 & 17.6 & 17.6 & 16.9 & 17.6 & 17.6 & \textit{17.6} & 23.7 & 23.7 & 23.1 & 23.7 & 23.7 & \textit{23.7} & 23.7 & 23.1 & 23,7 & 23.7 & \textit{23.7} \\
&&2&18.0 & 13.7 & 14.3 & 14.1 & \textbf{11.9} & \textit{10.7} & 20.3 & 19.6 & 23.2 & 23.3 & 18.6 & \textit{18.5} & 18.1 & 22.4 & 19.2 & \textbf{13.5} & \textit{11.0} \\
&&3&18.6 & 12.1 & 13.4 & 15.1 & \textbf{10.4} & \textit{8.5} & 23.8 & 24.5 & 25.7 & 36.3 & 24.6 & \textit{20.1} & 22.4 & 24.5 & 35.0 & \textbf{16.2} & \textit{14.0} \\
&&4&17.5 & 17.6 & 15.1 & 14.6 & \textbf{12.6} & \textit{10.2} & 27.6 & 31.2 & 28.7 & 41.7 & 35.1 & \textit{26.6} & 27.8 & 27.9 & 41.3 & \textbf{24.0} & \textit{19.6} \\
&&5&21.7 & 19.2 & 18.2 & 17.0 & \textbf{14.2} & \textit{11.6} & 28.4 & 30.3 & 31.3 & 50.7 & 36.9 & \textit{32.4} & 29.4 & 30.7 & 47.4 & \textbf{22.9} & \textit{19.6}\\
			
			\cline{2-20}
			
			&\multirow{5}{*}{\begin{tabular}[c]{@{}c@{}}Partially \\Changing \\ Classes\\ \end{tabular}}
			&1 & 17.6 & 17.6 & 16.9 & 17.6 & 17.6 & \textit{17.6} & 23.7 & 23.7 & 23.1 & 23.7 & 23.7 & \textit{23.7} & 23.7 & 23.1 & 23.7 & 23.7 & \textit{23.7} \\
&&2&17.4 & 16.0 & 15.0 & 14.5 & \textbf{13.6} & \textit{12.1} & 28.1 & 31.1 & 27.5 & 35.4 & 31.2 & \textit{26.7} & 29.9 & 26.9 & 28.2 & \textbf{25.8} & \textit{23.0} \\
&&3&17.1 & 14.9 & 15.8 & 14.2 & \textbf{13.2} & \textit{11.9} & 31.1 & 32.9 & 29.9 & 42.0 & 37.9 & \textit{29.4} & 32.1 & 29.1 & 39.2 & \textbf{27.6} & \textit{24.7} \\
&&4&13.5 & 12.8 & 13.4 & \textbf{10.5} & \textbf{10.8} & \textit{8.0} & 22.2 & 27.4 & 20.4 & 41.8 & 31.8 & \textit{22.1} & 25.8 & 20.5 & 36.9 & \textbf{18.8} & \textit{15.1} \\
&&5&16.9 & 14.9 & 16.1 & 12.2 & \textbf{9.9} & \textit{7.8} & 23.3 & 31.2 & 23.0 & 38.4 & 30.5 & \textit{25.5} & 28.2 & 23.3 & 34.4 & \textbf{19.6} & \textit{15.2}\\
			\hline \hline
			
		\parbox[t]{2mm}{\multirow{5}{*}{\rotatebox[origin=c]{90}{\textbf{Turbofan}}}} &
			
			\multirow{5}{*}{\begin{tabular}[c]{@{}c@{}}Not\\Applicable\\(Regression Task)\end{tabular}}
			&1 & 16.7 & 16.7 & 17.4 & 16.7 & 16.7 & \textit{16.7} & 17.3 & 17.3 & 18.2 & 17.3 & 17.3 & \textit{17.3} & 17.3 & 18.2 & 17.3 & 17.3 & \textit{17.3} \\
&&2&17.7 & 17.3 & 16.9 & 16.9 & \textbf{16.2} & \textit{15.7} & 18.7 & 18.2 & 19.0 & 18.9 & 18.4 & \textit{17.4} & 17.9 & 18.7 & 17.7 & \textbf{16.8} & \textit{15.9} \\
&&3&18.9 & 17.6 & 17.1 & \textbf{15.8} & 16.3 & \textit{15.0} & 19.3 & 18.7 & 22.9 & 19.3 & 17.6 & \textit{16.2} & 18.1 & 18.8 & 18.3 & \textbf{17.1} & \textit{15.8} \\
&&4&18.2 & 17.0 & 17.5 & \textbf{15.6} & \textbf{15.7} & \textit{14.7} & 16.5 & 17.4 & 17.9 & 17.9 & 17.5 & \textit{15.3} & 17.1 & \textbf{16.2} & 16.8 & \textbf{15.7} & \textit{15.1} \\
&&5&16.2 & 16.7 & 16.8 & 15.5 & \textbf{15.2} & \textit{13.9} & 17.5 & 17.6 & 18.2 & 18.3 & 17.2 & \textit{16.1} & 16.3 & 16.6 & 17.6 & \textbf{16.0} & \textit{14.9}\\
			\hline
	\end{tabular}}
	\label{tab:results}
\end{table*}

\begin{figure*}[h!]
   \captionsetup[subfigure]{font=small,labelfont=small, justification=centering}
    \centering

    \begin{subfigure}{0.23\textwidth}
        
        \centering
        \includegraphics[scale=0.39,trim={0.23cm 0.8cm 0.25cm 0},clip]{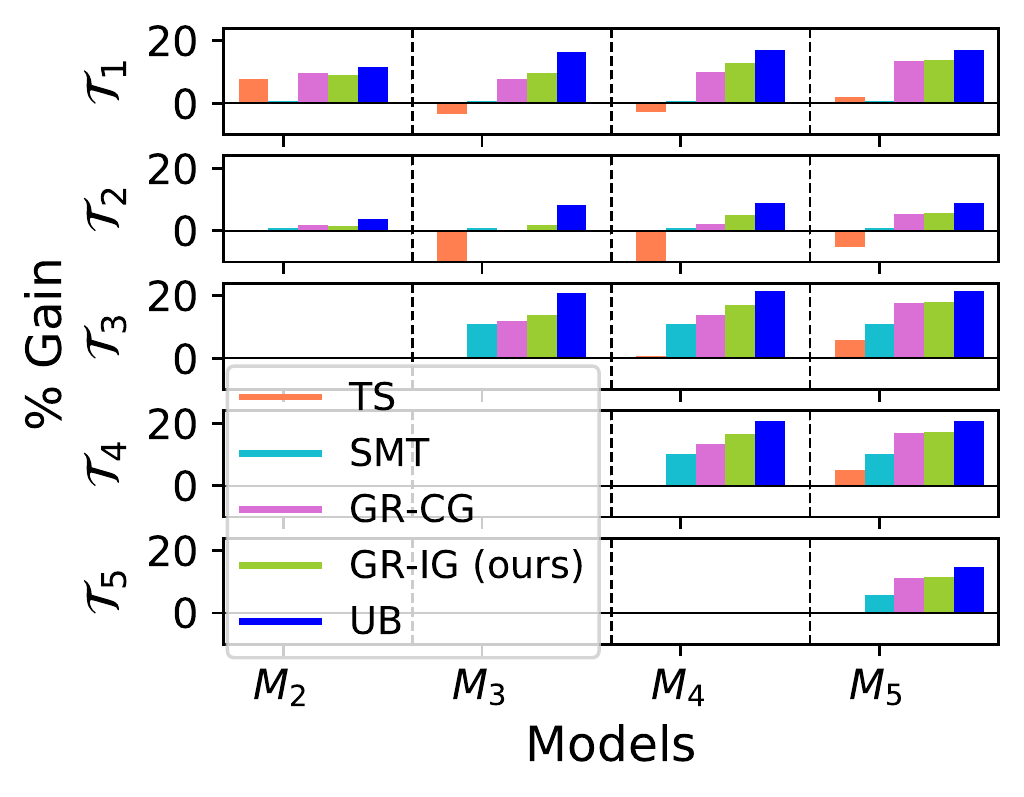} 
        \caption{DSADS \\(Fixed\\ Classes)}
    \end{subfigure}
    \begin{subfigure}{0.185\textwidth}
        \centering
        \includegraphics[scale=0.39,trim={2.15cm 0.8cm 0.25cm 0},clip]{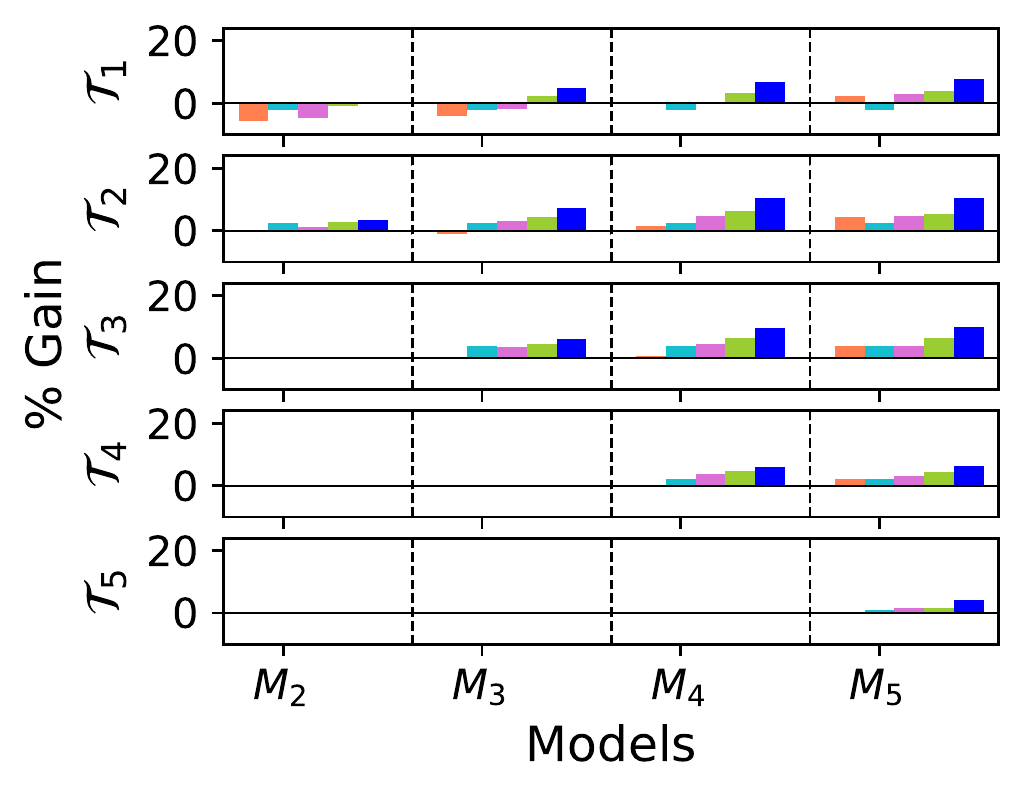} 
        \caption{DSADS \\(Incrementally \\Increasing Classes)}
    \end{subfigure}
    \begin{subfigure}{0.185\textwidth}
        \centering
        \includegraphics[scale=0.39,trim={2.15cm 0.8cm 0.25cm 0},clip]{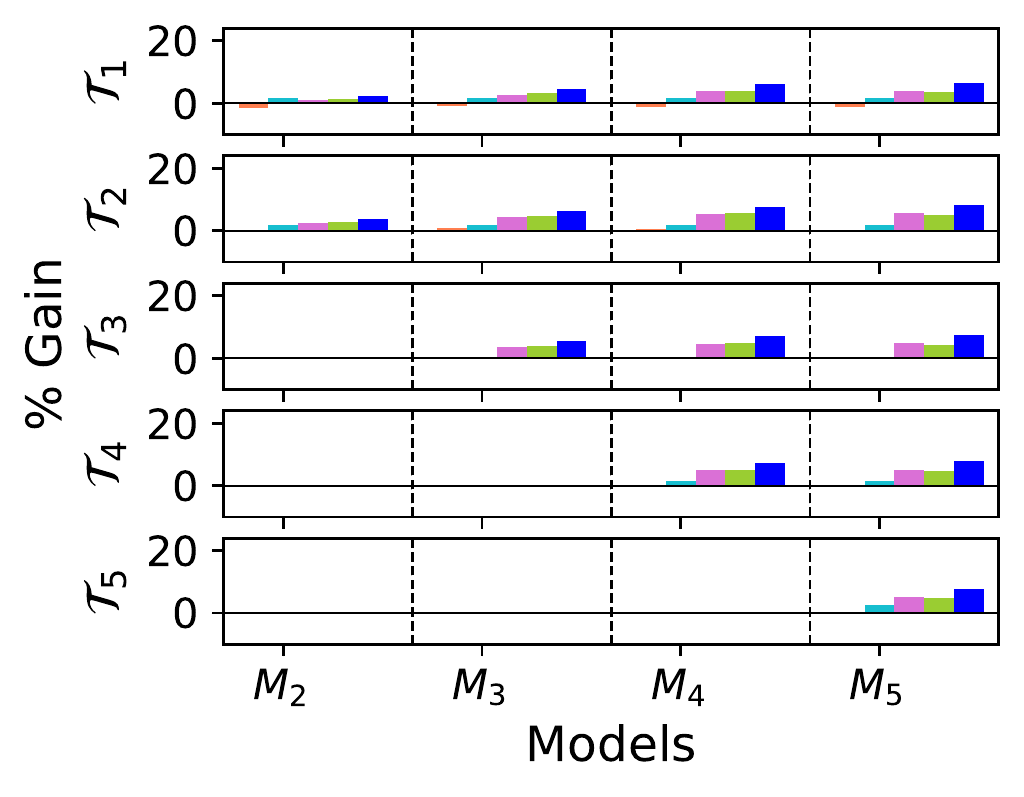} 
        \caption{HAR \\(Fixed \\Classes)}
    \end{subfigure}
    \begin{subfigure}{0.185\textwidth}
        
        \centering
        \includegraphics[scale=0.39,trim={2.15cm 0.8cm 0.25cm 0},clip]{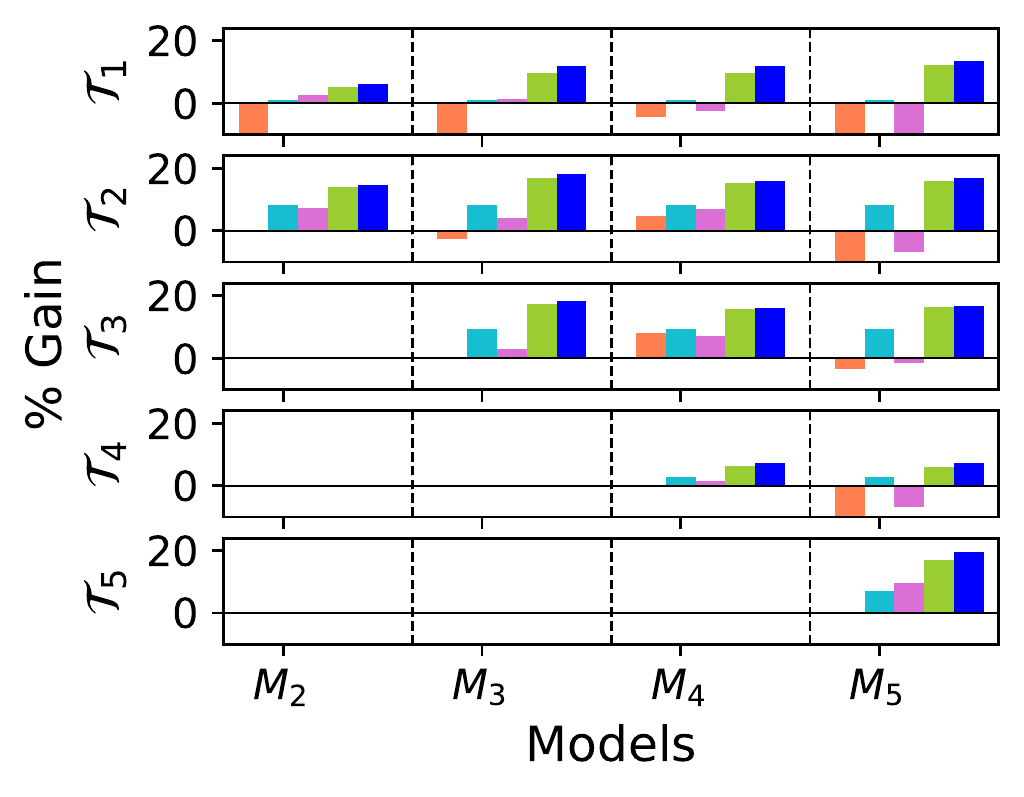} 
        \caption{HAR \\(Incrementally \\Increasing Classes)}
    \end{subfigure}
    \begin{subfigure}{0.185\textwidth}
        \centering
        \includegraphics[scale=0.39,trim={2.15cm 0.8cm 0.25cm 0},clip]{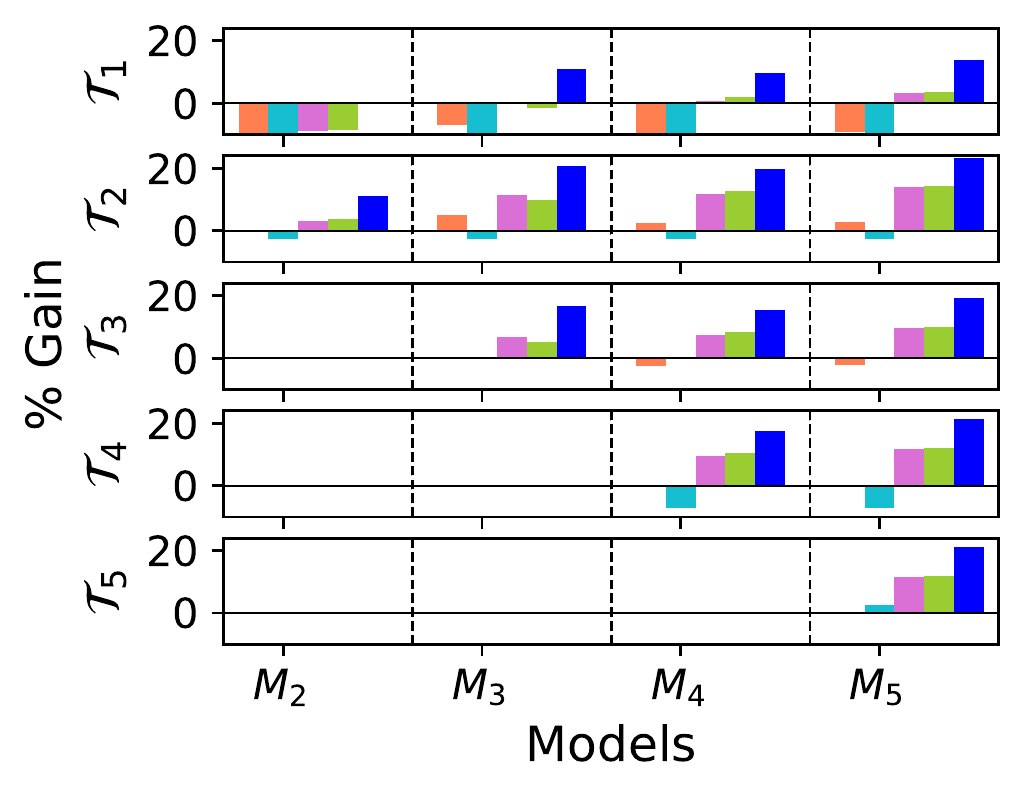} 
        \caption{Turbofan \\(Regression \\Task)}
    \end{subfigure}
    \vspace{2mm}
    
    \centerline{\newline(A) FID Scenario}
    \vspace{3mm}
    \begin{subfigure}{0.23\textwidth}
        \centering
        \includegraphics[scale=0.39,trim={0.23cm 0.8cm 0.25cm 0},clip]{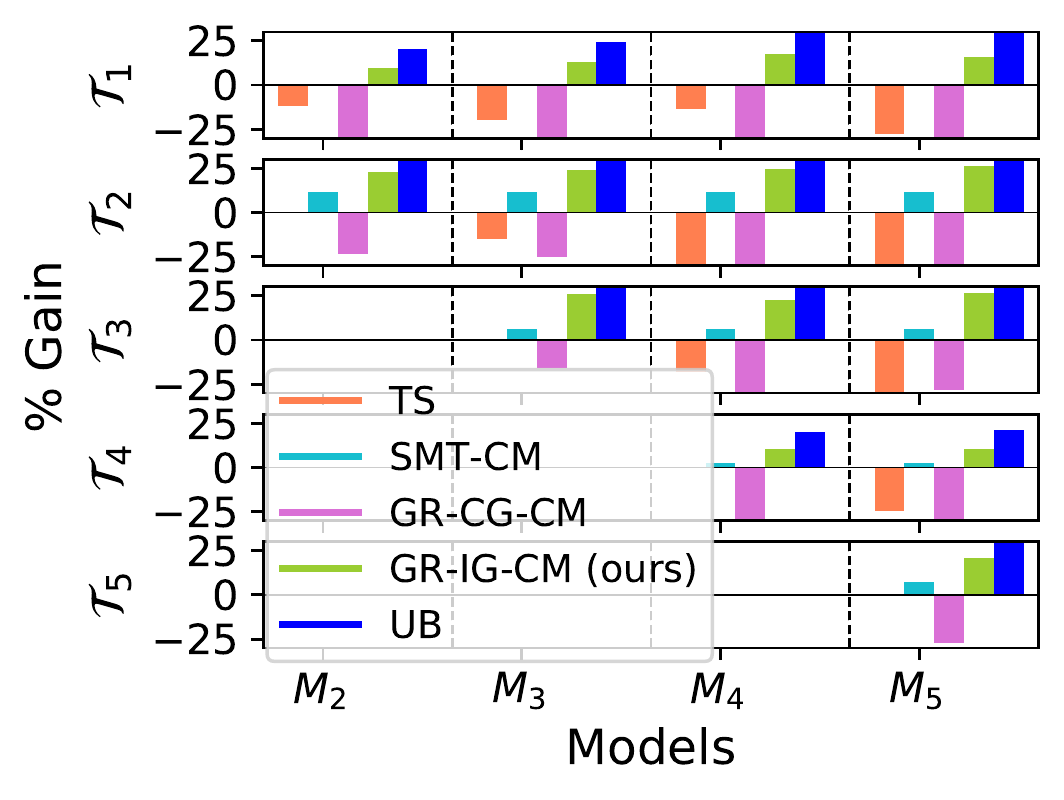} 
        \caption{DSADS \\(Fixed\\ Classes)}
    \end{subfigure}
    \begin{subfigure}{0.185\textwidth}
        \centering
        \includegraphics[scale=0.39,trim={2.45cm 0.8cm 0.25cm 0},clip]{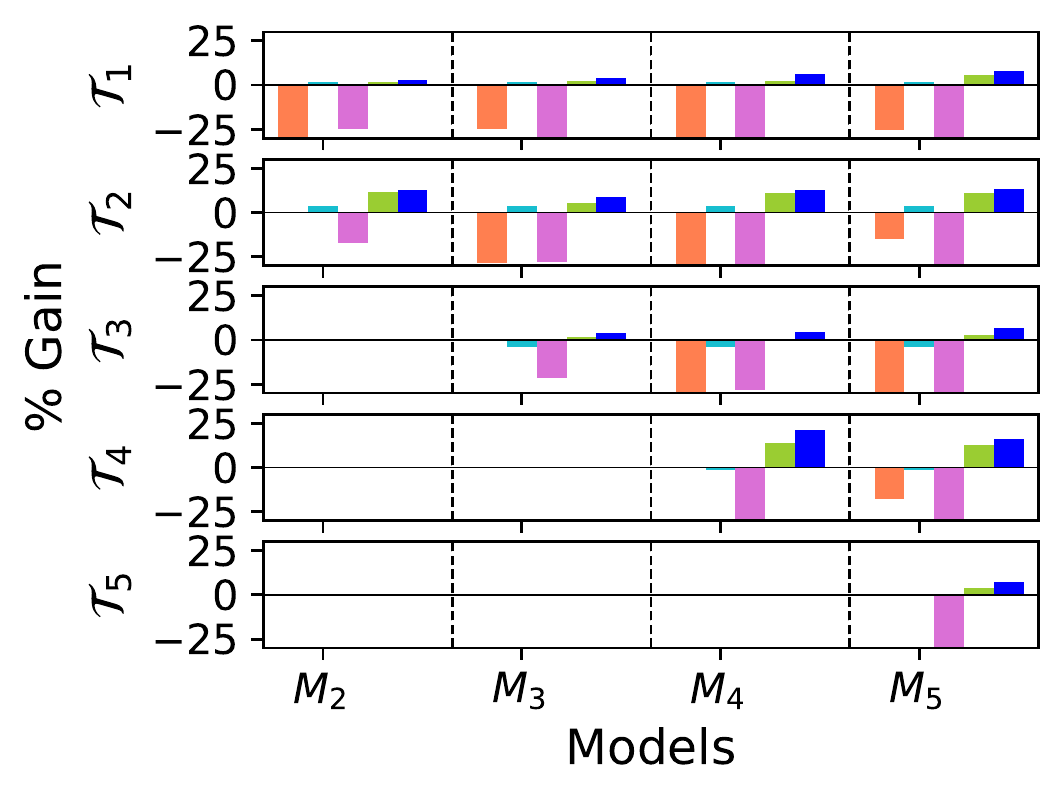}
        \caption{DSADS \\(Incrementally \\Increasing Classes)}
    \end{subfigure}
    \begin{subfigure}{0.185\textwidth}
        \centering
        \includegraphics[scale=0.39,trim={2.45cm 0.8cm 0.25cm 0},clip]{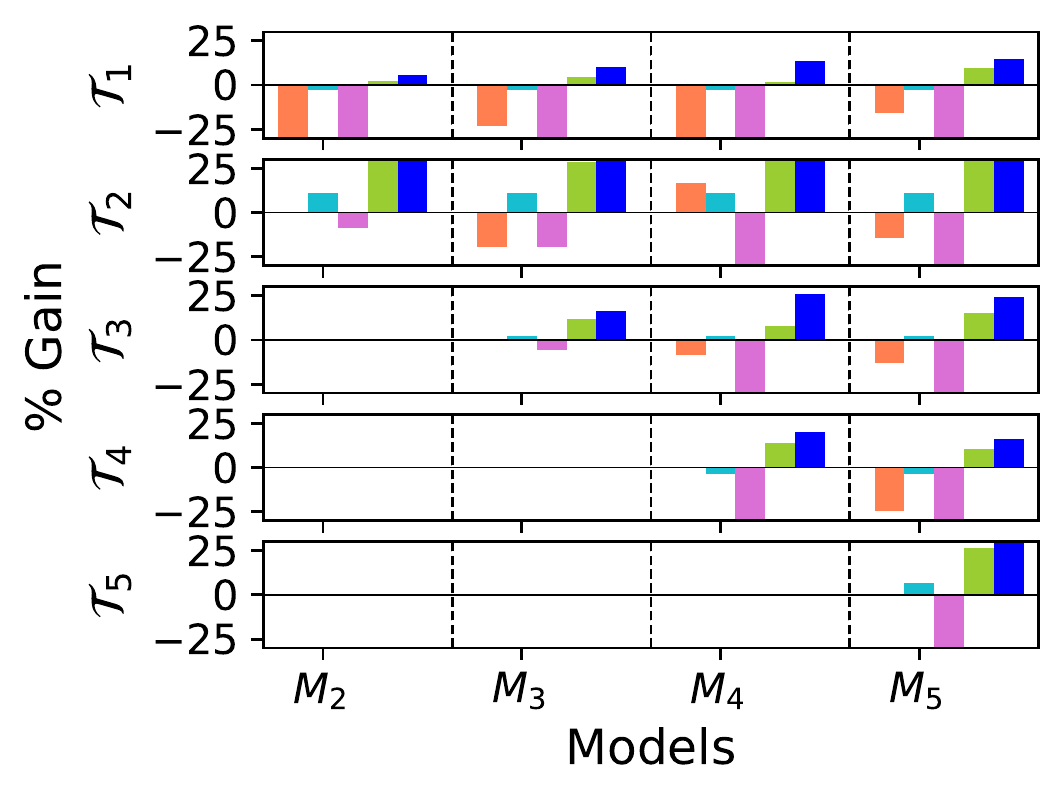} 
        \caption{HAR \\(Fixed \\Classes)}
    \end{subfigure}
    \begin{subfigure}{0.185\textwidth}
        
        \centering
        \includegraphics[scale=0.39,trim={2.45cm 0.8cm 0.25cm 0},clip]{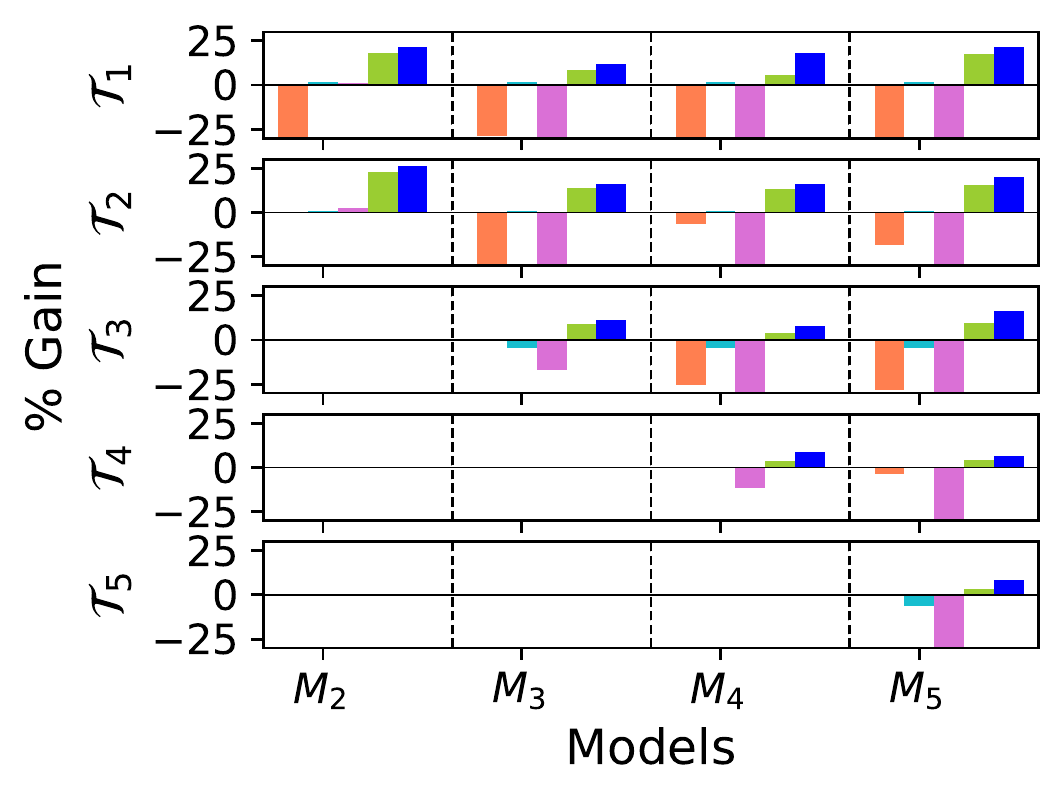} 
        \caption{HAR \\(Incrementally \\Increasing Classes)}
    \end{subfigure}
    \begin{subfigure}{0.185\textwidth}
        \centering
        \includegraphics[scale=0.39,trim={2.45cm 0.8cm 0.25cm 0},clip]{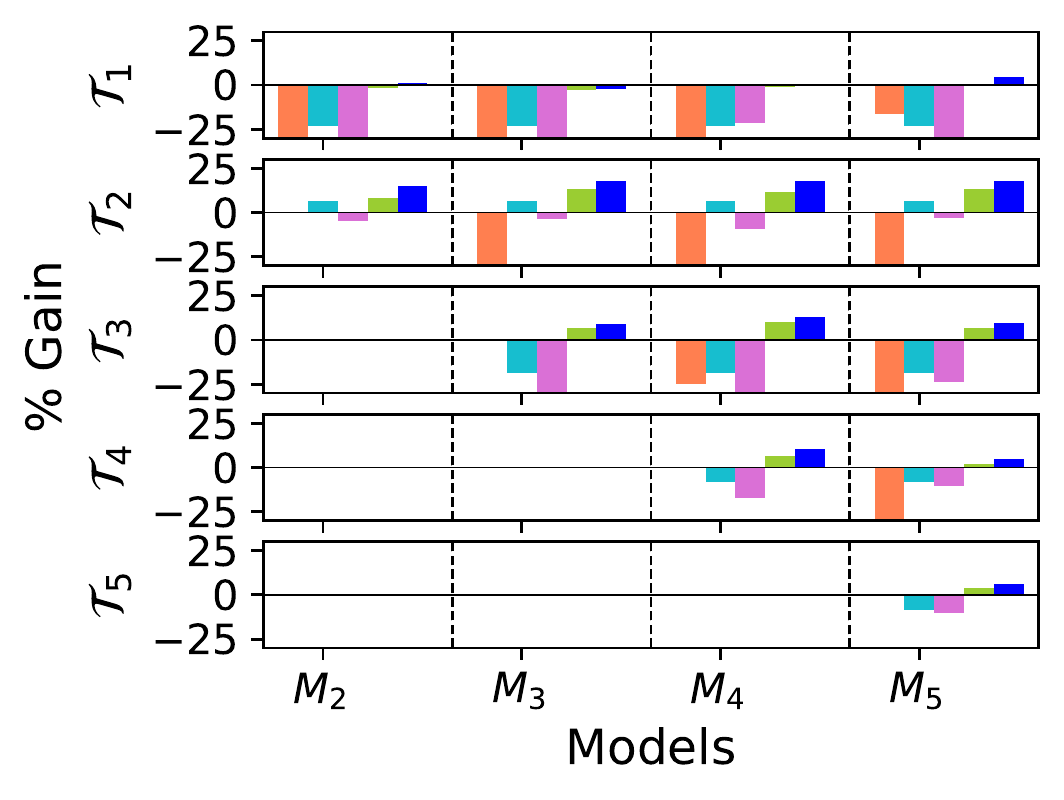} 
        \caption{Turbofan \\(Regression \\Task)}
    \end{subfigure}
    
    \vspace{2mm}
    
    \centerline{\newline(B) VID Scenario}
    \vspace{3mm}
    
    \caption{Percentage gain of respective solver model (Standard Solver in FID scenario and Solver-CM  in VID scenario) over the TS standard solver model. Respective solver model $M_i$ when used along with GR-IG have +ve gain on previous tasks $\mathcal{T}_{j}$ ($j=1,\ldots, i-1$) indicating the ability of the solver model with GR-IG to retain the knowledge of past tasks. TS standard solver model and solver model with GR-CG struggle to perform well on previous tasks, especially in the VID scenario. Our approach GR-IG is better than SMT indicating the advantage of generated previous tasks data from generative model and CM over SMT, which only uses the current task data to learn task-specific supermask. \label{fig:continual}}
\end{figure*}
\begin{enumerate}
    \item \textbf{Task-Specific (TS)} is the vanilla approach or baseline which uses only the data from current task, i.e. $\mathcal{D}_i$, for learning the task-specific model $M_i$. This approach serves as a lower bound for evaluation.
    \item \textbf{Fine-tune (FT)}: This approach also uses the data from the current task only (as in TS) but the parameters of the current solver model $M_i$ are initialized using those\footnote{For the final classification layer in case of incrementally increasing and partially changing classes, relevant parameters are used.} of the previous solver model $M_{i-1}$.
    \item \textbf{Common Generator (GR-CG)} \cite{shin2017continual}: This approach   maintains a single common generator (CG) across all previous tasks to generate samples from all of them (refer Section \ref{sec:approach}), while the solver model can be randomly initialized vanilla GRU model or the proposed conditioning module based solver model.
    \item \textbf{Supermasks with Transfer (SMT)} \cite{wortsman2020supermasks}: This approach maintains the same core solver model but learns task-specific supermasks. The task-specific supermask allows the model to retain knowledge to solve each task in different units of the same core base neural network. The supermask weights for the new task are initialised from the running mean of all the previous tasks supermask weights, resulting in transfer of knowledge. We found this initialization strategy to work better than the vanilla `Supermasks without Transfer' (SM) approach (in concurrence with the observations made in \cite{wortsman2020supermasks}), and report results for SMT alone.
	\item \textbf{Upper Bound (UB)}: This approach assumes access to all relevant original data from previous tasks. This approach is equivalent to having an ideal generator and solver for all previous tasks. Though access to original data from previous tasks is assumed, the available dimensions across tasks vary for the VID scenario while all dimensions are available for the FID scenario.
\end{enumerate}
Our proposed approach using independent generators for each previous task is referred to as \textbf{GR-IG}.
We refer to the GRU-based solver model using the GNN-based conditioning module as \textbf{Solver-CM}, while the vanilla GRU-based solver as \textbf{Standard Solver}.

\subsection{Results and Observations}
We make following key observations from Tables \ref{tab:results} and \ref{tab:Heterogeneousresults}:

    \begin{itemize}[leftmargin=*]
    \item As more tasks arrive, the gap between TS and UB widens, indicating scope for positive-transfer from old tasks to new tasks; this is expected as the underlying dynamical system is the same across tasks. We observe the same empirically: solvers using generative models (GR-CG and GR-IG) significantly improve upon TS in the FID scenario. For the partial changing classes scenario, GR-IG improves upon the baseline while GR-CG struggles, indicating the ability of GR-IG to provide better quality generated data. This is further helped by estimated labels from task-specific solvers, while GR-CG does not have this advantage.
	\item \textit{GR-IG performs better than GR-CG in most cases} with FID as well as VID, proving the advantage of GR-IG over GR-CG. The samples generated via GR-CG are the combination of all previous tasks with diversity across classes (and input dimensions in VID scenario), which cannot be captured by one VAE. 
	\item SMT performs better than TS on FID scenario with standard solver whereas, on VID scenario SMT with Standard Solver performs worse than TS mostly in incrementally increasing classes and partial changing classes. The gains observed when using Solver-CM instead of Standard Solver for VID scenario confirms the advantage of GNN-based conditioning module for dimension-adaptability.  
 Our proposed approach always performs better than SMT in both FID and VID scenarios, showing the advantage of using the independent generator (GR-IG) for each task.
	\item For scenarios with partially-observed sensors, i.e. VID scenario, GR-CG performs worse than TS. Unlike the FID scenario, even GR-IG and UB with access to all data from previous tasks fail to improve upon TS when using standard solver.
	Furthermore, vanilla transfer via FT models is ineffective in VID scenario.
	These observations confirm the \textit{non-trivial nature of the VID problem}, and inability of GR-CG to adapt to variable dimensions.
	\textit{Only Solver-CM with GNN-based Conditioning Module for dimension-adaptability and GR-IG for generating past data consistently performs better than the TS on VID scenarios}. Combination of Solver-CM and GR-IG is the best performing model in most scenarios across datasets and tasks.
	\item \textit{Continual learning}: 
	We also evaluate how well Solver-CM  in VID scenario and standard solver in FID scenario with GR-IG performs on tasks $\mathcal{T}_1,\ldots,\mathcal{T}_{i-1}$. As shown in Fig. \ref{fig:continual}, in both FID and VID scenarios, we observe that while the TS model on one task struggles to generalize on other tasks, our approach performs better than SMT and GR-CG and is comparable to the UB in all scenarios.  This shows that the \textit{samples generated from GR-IG are better than GR-CG in both FID and VID scenarios and also solvers with CM (GNN-based conditioning module) are able to adapt variable dimension in VID scenario}, which in-turn helps to solve the previous tasks better by virtue of capabilities similar to generative replay \textit{GR-CG} as in \cite{shin2017continual}.
	\begin{table}[ht]
	\centering
	\caption{Performance comparison for generalization to unseen sensor combinations at test time.   \textit{Solver-MH}: the conditioning module (CM) is removed and the conditioning vector is replaced with a multi-hot vector indicating the active sensors. \textit{Solver-SE}: the GNN is removed and replaced with a simple max-pooling over the sensor embeddings. Standard Solver-A: the upper bound that assumes access to all sensors in the train and test set. \textit{In both the settings, No Fine-tuning and Fine-tuning, Solver-CM generalizes better on unseen sensor combination than other solver models.} }
	\scalebox{0.82}{
		\begin{tabular}{|c|cccc|cccc|c|}
			\hline
			&\multicolumn{4}{c}{\textbf{No Fine-tuning}} & 
			\multicolumn{4}{|c|}{\textbf{Fine-tuning}} & \\
			\cline{2-9}
			&\multicolumn{8}{c|}{\textbf{Solver-}}& \\
			\cline{2-9}
			{\begin{tabular}[c]{@{}c@{}}\textbf{Dataset}\\\\\end{tabular}} 
            &{\begin{tabular}[c]{@{}c@{}}\textbf{Stan-}\\\textbf{dard}\end{tabular}} &{\begin{tabular}[c]{@{}c@{}}\textbf{
            MH}\end{tabular}}& {\begin{tabular}[c]{@{}c@{}}\textbf{SE}\end{tabular}} & {\begin{tabular}[c]{@{}c@{}}\textbf{CM}\end{tabular}} & {\begin{tabular}[c]{@{}c@{}}\textbf{Stan-}\\\textbf{dard}\end{tabular}}  &{\begin{tabular}[c]{@{}c@{}}\textbf{MH}\end{tabular}}& {\begin{tabular}[c]{@{}c@{}}\textbf{SE}\end{tabular}}& {\begin{tabular}[c]{@{}c@{}}\textbf{CM}\end{tabular}} & {\begin{tabular}[c]{@{}c@{}}\textbf{Standard}\\\textbf{Solver-A}\\\textbf{(UB)}\end{tabular}} \\ 
            \hline
			DSADS  
			&  8.3  &8.9&7.8& \textbf{5.9} & 4.7 &6.1& 3.4 & \textbf{3.3} & 1.5 \\  
			\hline
			HAR  
			& 12.8& 13.5  &11.8& \textbf{11.4} & 12.1 & 12.4&10.7 & \textbf{10.5} &  6.3  \\
			\hline
			Turbofan
			& 15.5  & 16.4&14.9&\textbf{14.6} &  16.2 &15.6&14.7 & \textbf{14.2} &13.6 \\
			\hline
		\end{tabular}
	}
	\label{tab:Heterogeneousresults}
\end{table}
	\item \textit{Ablation Study of Solver-CM}: We consider VID multi-sensor tasks with 100\% training instances instead of the scarce training instances in the sequential-tasks setup discussed above. The training time-series correspond to certain combinations of sensors, while the test set contains sensor combinations unseen during training.
	Any time-series instance in train or test set has $\approx 0.4$ fraction sensors missing. For all datasets, 40\% of the instances are used for training, 10\% for validation, 10\% for fine-tuning (ignored in no fine-tuning setting), and remaining 40\% for testing. From Table \ref{tab:Heterogeneousresults} we observe that Solver-SE is better than Solver-MH, indicating the advantage of using the compressed information of active sensors in the form of sensor embeddings over a multi-hot vector. But still, Solver-SE fails to generalize on unseen sensors combinations. We use GNN and sensor embeddings in Solver-CM to improve further, where GNN allows to exchange messages between the active sensors. These messages passing amongst the active sensors inherently allow for combinatorial generalization on unseen sensor combinations without fine-tuning. For fine-tuning, we consider small amounts of data for the same sensor combinations as the test set, which further improves the results.
	
 \end{itemize}
In summary, while GR-CG class of methods like \cite{shin2017continual,velik2014brain} would usually suffice to deal with lack of access to past data by providing generated data, and SMT \cite{wortsman2020supermasks} would allow to adapt across tasks with varying classes via supermasks or similar methods, neither class of methods suffice for the variable input dimensions scenario as indicated by their inferior performance in comparison to GR-IG with CM-based solver. Furthermore, conditioning module based on GNNs is critical to the success of the proposed approach.

\section{Conclusion and Future work}
In this work, we have motivated a practical problem of enabling knowledge transfer from previous tasks to new tasks under restricted data-sharing across sequentially arriving multi-sensor tasks with varying input dimensions. We have proposed a deep learning architecture to deal with
multivariate time series classification and regression tasks that works across the partially-observed sensor (variable input dimensions) tasks as well as changing classes across tasks.
Our results on three publicly available datasets prove the efficacy of the proposed approach to leverage generated data from previous tasks to improve performance on (partially-observed multi-sensor) new tasks, without any explicit original data exchange.
Though we have evaluated the proposed approach for activity recognition and remaining useful life estimation applications, we believe the proposed approach is generic and would be useful in other multivariate time series applications. 
Furthermore, combining the benefits of generative replay as we have presented, with meta-learning (e.g., on lines of \cite{riemer2018learning})
to further improve transfer of the solver models used in our approach is an interesting future direction to explore
especially in context of variable-dimensional input spaces. Extensions of this work in privacy-preserving scenarios is another interesting future direction.
\bibliographystyle{abbrv}
\bibliography{BibTex/sensor_analytics,BibTex/ijcai20,BibTex/aaai,BibTex/dataset}

\appendix

\section{Approach}\label{sup: Approach Appendix}
\subsection{Generator Models}\label{sup: Generator Models}
We use RNN-based variational autoencoders (VAEs) generative model to generate previous tasks training instances. The detail description of VAEs are as follows:
VAEs consist of two main parts: Encoder that learns a mapping from data to the latent representations, Decoder from latent representations to generate data. Autoencoders compress the large input features into smaller representation which can latter be reconstructed by decoder. The input time-series of task $i$ are processed by encoder $(E_i)$ of generator $(G_i)$, where $\boldsymbol{\theta}_{E_i}^{GRU}$ are the learnable parameters of GRU-based RNN as follows:
\begin{align}
p(\mathbf{z}_{i, r}| \mathbf{x}_{i, r}, \boldsymbol{\theta}_{E_i}^{GRU}) = GRU(\mathbf{x}_{i, r}, \boldsymbol{\theta}_{E_i}^{GRU}).
\end{align}
The output of encoder is mean and log-variance vectors (mean $\mu$ and log-variance $\sigma^2$ are learnable parameters of the network). Then the reparameterization trick \cite{kingma2013auto} is done to update the network's parameters of VAE using back-propagation. 
\begin{align}
\mathbf{z}_{i, r} = \boldsymbol{\mu}_{i, r} + \boldsymbol{\sigma}_{i, r}^2 \odot \boldsymbol{\epsilon}, \quad \boldsymbol{\epsilon} \sim \mathcal{N}(0, I).
\end{align}
Decoder $(D_i)$ of generator $(G_i)$ then takes sample $\mathbf{z}_{i, r} \sim p(\mathbf{z}_{i, r}| \mathbf{x}_{i, r}, \boldsymbol{\theta}_{D_i}^{GRU})$ to reconstruct time-series.
\begin{align}
q(\hat{\mathbf{x}}_{i, r}| \mathbf{z}_{i, r}, \boldsymbol{\phi}_{D_i}^{GRU}) = GRU(\mathbf{z}_{i, r}, \boldsymbol{\phi}_{D_i}^{GRU}).
\end{align}
\subsubsection{Generator Training}
The objective is to minimize the following loss over all training samples:
\begin{dmath}\label{generator}
\mathcal{L}_{\text{VAE}} (\mathbf{x}_{i, r}) = {\frac{1}{2T_i}{\sum_{t=1}^{T_i}||\mathbf{x}_{i, r}^t - \hat{\mathbf{x}}_{i, r}^t||^2} } + {\quad \frac{\beta}{2d'}{\sum_{j= 1}^{d'}\big( \boldsymbol{\mu}_{j}^{2} + \boldsymbol{\sigma}_{j}^{2} - 1 - log(\boldsymbol{\sigma}_{j}^{2}) \big)}},
\end{dmath}

where $||.||$ denotes the standard $L_2$ norm, $d'$ denotes the dimension of the latent representations, 
the first term is the reconstruction loss and the second term corresponds to the Kullback-Leibler (KL) divergence loss w.r.t. $\mathcal{N}(\boldsymbol{0},\boldsymbol{I})$ (refer \cite{kingma2013auto} for more details), $\beta$ denotes contribution of KL divergence towards loss function.
Once the generator $G_i$ of task $\mathcal{T}_i$ is trained, the learned decoder $D_i$ takes samples from $z_{r} \in \mathcal{N}(0, I)$ to reconstruct time-series $\hat{\mathbf{x}}_{i, r}$ which is defined as: $\hat{\mathbf{x}}_{i, r} = D_{i}(z_{r}; \boldsymbol{\phi}_{D_i}^{GRU})$.

The loss functions of common generator (GR-CG) and independent generator (GR-IG) are as follows:

\begin{figure*}[h!]
	\centering
	\includegraphics[scale=0.18,trim={0cm 19cm 4cm 10.5cm}, clip]{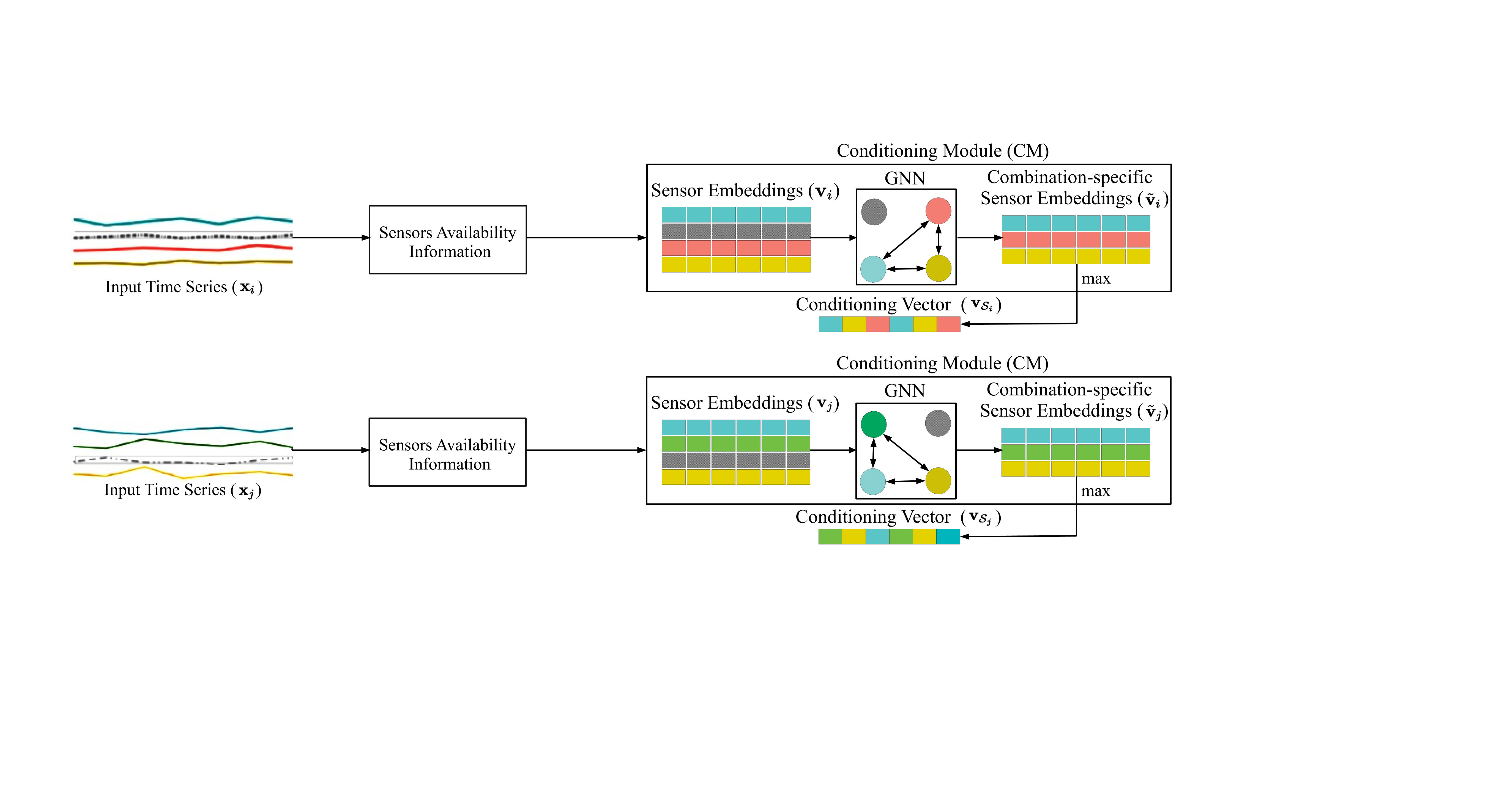}
	
	\caption{There are two multivariate time series $\mathbf{x}_i$ and $\mathbf{x}_j$ which have different subset of available sensors. Thus, the conditioning module generate a conditioning vector, specific to combination of available sensors in a particular input time series. \label{fig:Conditioning module}}
\end{figure*}
\subsubsection{Common Generator} In common generator, a single generator is maintained across tasks. Input to common generator for task $i$ are samples of current task $\mathbf{x}_{i, r}$ and the generated samples of all previous task $\hat{\mathbf{x}}_{j, r}$ $(j = 1, \ldots, i-1)$, which are generated by the previous generator of task $i-1$. The current task samples and all the previous generated tasks samples are mixed in a ratio $q$, i.e. weightage given to current task samples as compared to all previous tasks generated samples. Thus, the trained generator for task $i$ will generate samples combination of current task $i$ and all previous tasks $1, \ldots, i-1$. Note: Decoder $D_{i-1}$ of the learned generator $G_{i-1}$ is only used to generate all previous tasks $\mathcal{T}_j (j = 1, \ldots, i-1)$ samples. The loss function of common generator for task $i$ is defined as: 

\begin{dmath}\label{CG}
	\mathcal{L}_{G_i} ={ q(\mathcal{L}_{VAE} (\mathbf{x}_{i, r}))} + {(1-q)(\mathcal{L}_{VAE}(D_{i-1}(z_{r}; \boldsymbol{\phi}_{D_{i-1}}^{GRU})))},
\end{dmath}

where, $0< q \leq 1$ is the importance given to samples of current task.

\subsubsection{Independent Generator}
In independent generator, a separate generator is maintained for each task. The loss function of independent generator is similar to the common generator, except the $q$ value is set to 1 in above Eqn. \ref{CG}.

\subsection{Solver Model}
\subsubsection{Training Objective}
We use the standard cross-entropy and squared-error losses as training objectives for the classification and regression tasks, respectively.

\begin{dmath}
	\mathcal{L}(y_{i,r}, \hat{y}_{i, r}; \boldsymbol{\theta}_{i})= \left \{
	\begin{aligned}
		&-\sum_{k=1}^{K}{y_{i, r}^k log(\hat{y}_{i, r}^k}) && \text{for classification tasks} \\
		& (y_{i,r}-\hat{y}_{i,r})^2 && \text{for regression tasks}
	\end{aligned} \right.
\end{dmath} 
where, $K$ denotes number of classes.

\section{EVALUATION}
\subsection{Dataset Details}\label{sup:dataset}
We perform extensive evaluations on two publicly available human activity recognition benchmark datasets\footnote{\url{https://github.com/titu1994/MLSTM-FCN/releases}} used by \cite{karim2019multivariate} and an RUL estimation Turbofan Engine dataset\footnote{\url{https://ti.arc.nasa.gov/tech/dash/groups/pcoe/prognostic-data-repository/##turbofan}}: 
\begin{enumerate}
    \item \textbf{Daily and Sports Activities Data Set (DSADS)} \cite{altun2010human}: In DSADS each activity is recorded using sensors installed on different body parts of 4 female and 4 male.
    \item \textbf{Human Activity Recognition Using Smartphones (HAR)} \cite{anguita2012human}: In HAR, 30 participants performed daily living
activities by wearing a waist-mounted smartphone with embedded inertial sensors.

    \item \textbf{Turbofan Engine} \cite{saxena2008damage}: We have used FD001 of the turbofan engines dataset repository containing time-series of readings 14 sensors without any operating condition variables. The sensor readings for the engines in the training set are available from the starting of their operational life till the end of life or failure point, while those in the test set are clipped at a random time prior to the failure, and the goal is to estimate the RUL for these test engines. 
\end{enumerate}
\begin{table}[h]
\caption{Dataset details. Here, $d$: maximum available sensors in a time-series, C: classification, R: regression, $N$: number of instances, $K$: number of classes.}
\centering
	\scalebox{1}{
		\centering
		\begin{tabular}{|l|c|c|c|c|c|c|}
			\hline
			{Dataset} & $d$ & {Task} & \textbf{$N$} & $K$ \\
			\hline
			DSADS & 45 & C & 9,120 & 19 \\
			HAR & 9 & C & 10,299 & 6  \\
			Turbofan - FD001 & 14 & R & 200 & - \\
			\hline
	\end{tabular}}
	
	\label{tab:datasets}
\end{table} 
\subsection{Hyperparameters Used}\label{sup:para}
The encoder of generator is modeled by three GRU layers, each with 128, 256 and 64 units in DSADS, HAR and Turbofan, respectively. The encoder compresses the input into latent space representations using a linear layer with 40, 80 and 40 units in DSADS, HAR and Turbofan, respectively, to output a mean vector and a standard deviation. The architecture of the decoder is pretty much the reverse of the encoder.  We use Adam optimizer \cite{kingma2015adam} with a learning rate of $1e-4$ and epoch 500 for training generator from scratch. The value of the mixing ratio $q$ is 0.5 in the case of common generator, whereas in independent generator value of $q$ is 1.

For classification tasks, CDM consists of three GRU layers with 128 units each and for regression tasks, CDM consists of three GRU layers with 60 units each. We use a mini-batch size of 32, and $d'$=$\lfloor\frac{d}{2}\rfloor$ for the combination-specific conditioning vector of $\mathbf{v}_{\mathcal{T}_{i}}$ during training.

All feedforward layers are followed by dropout \cite{srivastava2014dropout} of 0.2 for regularization in addition to early stopping with a maximum of 250 epochs for training.
We use vanilla SGD optimizer without momentum to update the sensor embedding vectors with a learning rate of $5e-4$, and Adam optimizer to update the rest of the layers with a learning rate of $1e-4$. Since the active nodes change with changing combinations of the available sensors, we found it useful to use vanilla SGD for updating the sensor vectors (else, if we use momentum, the vectors for the inactive nodes would also get updated). On the other hand, the GNN and the CDM are shared across all combinations and mini-batches benefit from momentum, and hence Adam is used for updating their parameters.
\end{document}